\documentclass[twocolumn,10pt]{article}
\usepackage[a4paper,margin=1in]{geometry}
\usepackage{cite}
\usepackage{amsmath,amssymb,amsfonts}
\usepackage{algorithmic}
\usepackage{graphicx}
\graphicspath{{images/}}
\usepackage{booktabs}
\usepackage{pifont}
\usepackage{textcomp}
\usepackage{mathptmx} 
\usepackage{tabularx}
\usepackage{multirow}
\usepackage{hyperref}
\usepackage{cleveref}
\usepackage{longtable}
\usepackage{ltablex}
\keepXColumns
\usepackage{soul}
\renewcommand{\hl}[1]{#1}
\usepackage{float}
\usepackage{bm}
\usepackage{colortbl}
\usepackage{xcolor}
\usepackage{newtxtext,newtxmath}
\usepackage{cuted}   
\usepackage{stfloats} 
\usepackage{placeins} 
\definecolor{YellowOrange}{RGB}{255, 174, 66} 
\definecolor{cyan20}{RGB}{204, 255, 255}
\definecolor{red20}{RGB}{255, 204, 204}
\definecolor{darkRed}{RGB}{189, 0, 0}
\definecolor{darkBlue}{RGB}{0, 0, 189}
\definecolor{darkGreen}{RGB}{0, 140, 0}
\definecolor{PeaGreen}{RGB}{160,220,160}
\definecolor{none}{RGB}{214,220,211}
\definecolor{samrgb}{RGB}{233,203,226}
\definecolor{samlidar}{RGB}{184,226,233}
\definecolor{samrgblidar}{RGB}{239,209,181}
\definecolor{SAM}{RGB}{206, 206, 253}
\definecolor{Inj}{RGB}{128, 181, 104}
\definecolor{Ext}{RGB}{97, 203, 244}
\definecolor{mod}{RGB}{251, 229, 133}
\definecolor{modrgb}{RGB}{251, 229, 133}
\definecolor{modother}{RGB}{217, 205, 168}
\definecolor{fusion}{RGB}{252, 182, 177}
\definecolor{seghead}{RGB}{166, 191, 215}

\usepackage{authblk}
\title{\textbf{Multimodal SAM-adapter for Semantic Segmentation}}
\author[1]{Iacopo Curti}
\author[1]{Pierluigi Zama Ramirez}
\author[2]{Alioscia Petrelli}
\author[1]{Luigi Di Stefano}
\affil[1]{University of Bologna, Bologna 40136, Italy}
\affil[2]{SINA, Bologna 40132, Italy}
\date{} 

\begin{document}
\maketitle

\begin{strip}
\begin{abstract}
Semantic segmentation, a key task in computer vision with broad applications in autonomous driving, medical imaging, and robotics, has advanced substantially with deep learning. Nevertheless, current approaches remain vulnerable to challenging conditions such as poor lighting, occlusions, and adverse weather. To address these limitations, multimodal methods that integrate auxiliary sensor data (e.g., LiDAR, infrared) have recently emerged, providing complementary information that enhances robustness.  
In this work, we present MM SAM-adapter, a novel framework that extends the capabilities of the Segment Anything Model (SAM) for multimodal semantic segmentation. The proposed method employs an adapter network that injects fused multimodal features into SAM’s rich RGB features. This design enables the model to retain the strong generalization ability of RGB features while selectively incorporating auxiliary modalities only when they contribute additional cues. As a result, MM SAM-adapter achieves a balanced and efficient use of multimodal information.  
We evaluate our approach on three challenging benchmarks, DeLiVER, FMB, and MUSES, where MM SAM-adapter delivers \textit{state-of-the-art} performance. To further analyze modality contributions, we partition DeLiVER and FMB into \textit{RGB-easy} and \textit{RGB-hard} subsets. Results consistently demonstrate that our framework outperforms competing methods in both favorable and adverse conditions, highlighting the effectiveness of multimodal adaptation for robust scene understanding.\\
The code is available at the following \href{https://github.com/iacopo97/Multimodal-SAM-Adapter}{GitHub Repository}.
\end{abstract}
\paragraph{Keywords}
adapter, event cameras, thermal cameras, depth, LiDAR, multimodal semantic segmentation, SAM
\end{strip}
\FloatBarrier

\section{Introduction}
\label{sec:introduction}

Semantic segmentation is a fundamental computer vision task that assigns a category label to each image pixel, with applications in fields such as autonomous driving, medical image analysis, and robotic navigation. The advent of deep learning has revolutionized semantic segmentation, leading to remarkable improvements in accuracy, efficiency, and generalization across environments \cite{thisanke2023semantic}.
Despite advances, semantic segmentation from RGB images fails in challenging situations such as poorly lit scenes, adverse weather conditions, and motion blur. To overcome these limitations, additional sensors that provide complementary measurements, such as infrared or event cameras and LiDAR,  may be deployed to enhance segmentation performance.
\begin{figure}[!h]
    \centering
    \includegraphics[width=0.99\linewidth]{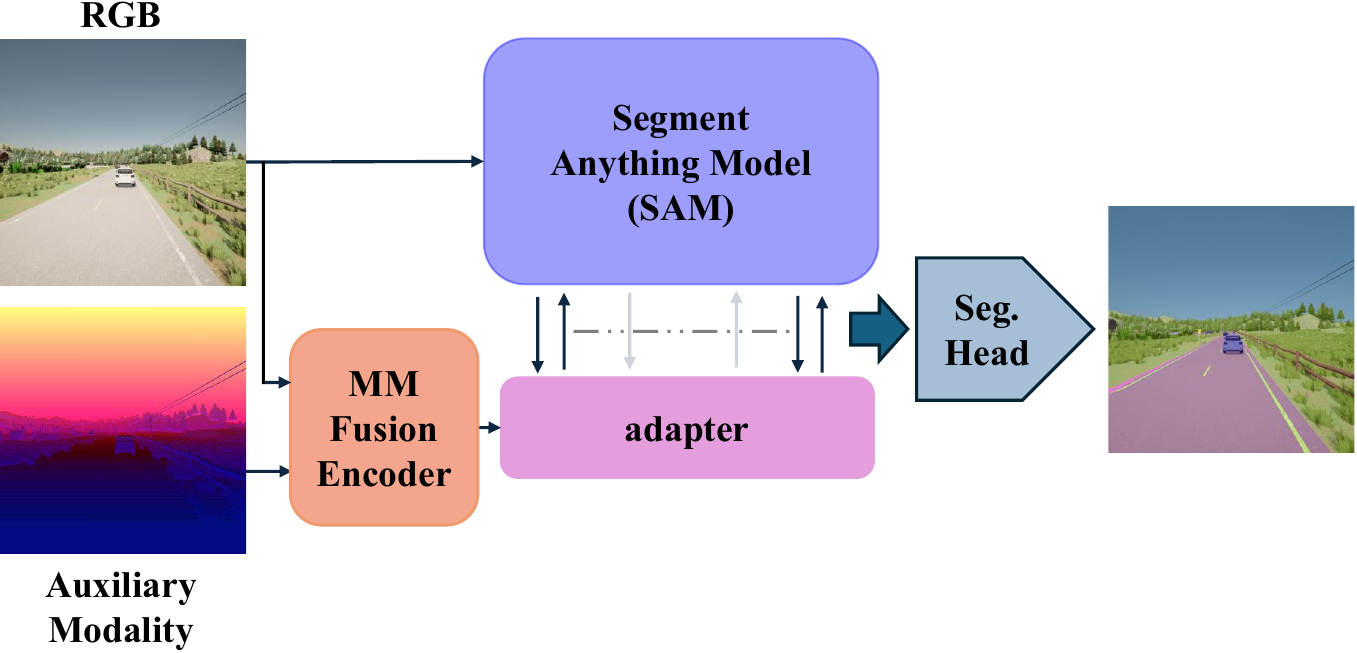}
    \caption{We propose adapting the Segment Anything Model's rich and general knowledge for multimodal semantic segmentation using an external adapter network. Thanks to a Multimodal Fusion Encoder, our adapter leverages both the RGB and auxiliary modality (depth map in the Figure) information to achieve optimal performance on both challenging and easy scenarios.}
    \label{fig:teaser_network}
\end{figure}
The increasing adoption of auxiliary sensors has fostered the creation of numerous datasets and approaches \cite{zhang2021deep} that use multiple input modalities to pursue semantic segmentation.
Standard practice for state-of-the-art multimodal methods is to train networks on the target dataset starting from ImageNet \cite{5206848} weights, i.e., building upon the foundational knowledge acquired by image classification.

However, the recent availability of large-scale annotated datasets and large GPU clusters has enabled the development of foundational segmentation models capable of accurately segmenting any RGB image. Among these, the Segment Anything Model (SAM) \cite{Kirillov_2023_ICCV}, trained on 11 million images and 1 billion ground truth masks, stands out for its impressive generalization capabilities.
The availability of such strong foundational models for segmentation made us question whether we could employ the segmentation-specific knowledge embedded in models like SAM to tackle multimodal semantic segmentation. Although SAM is designed to perform instance-level segmentation via point or box prompts on RGB images, we conjecture that the fine-grained features dealing with spatial and contextual relationships between objects and parts extracted by the SAM encoder may be used to address pixel-level semantic segmentation even when processing RGB images alongside other modalities. Thus, in this paper, we propose to deploy the rich and generalizable features learned by SAM to achieve multimodal semantic segmentation. In particular, our goal is to adapt SAM to multimodal semantic segmentation while retaining its previous general knowledge.
To achieve our goal, we draw inspiration from the strategy proposed by ViT-adapter \cite{chen2023vision}, which pertains to tailoring a pre-trained Vision Transformer (ViT) \cite{vit} for dense tasks by an external adapter network. Based on the mechanism of cross-attention, this design can adapt the foundational ViT \cite{vit} features learned through image classification to tasks such as semantic segmentation, avoiding catastrophic forgetting. In our work, we employ a similar external adapter (the pink block in \Cref{fig:teaser_network}), yet our goal is not only to adapt SAM features to a new task, namely from instance segmentation via prompts to semantic segmentation,  but also to integrate features extracted from an auxiliary modality, such as, e.g., depth maps, LiDAR measurements, thermal or event data. 
Our architecture combines SAM alongside an adapter network. This design follows our intuitions that the RGB features learned by SAM through massive pre-training on billions of images can serve as a strong foundation for multimodal semantic segmentation and that tailoring to the new task without 
forgetting reusable knowledge about instances and their parts may be achieved effectively through modern adaptation strategies such as \cite{chen2023vision}.
To prioritize its general and rich RGB features, we combine SAM with a lighter adapter network geared to incorporate the auxiliary modality. This asymmetric design materializes our intuition that, due to the outstanding segmentation capabilities of image-based models like SAM, multimodal semantic segmentation may best be tackled by relying mainly on RGB images and utilizing other modalities to handle challenging situations where the image turns out insufficiently informative. This starkly contrasts traditional multimodal architectures, which typically include networks of similar capacity to process the different modalities, thereby implicitly weighing each modality equally.
Our experiments show that the SAM adaptation strategy outlined so far can handle challenging settings effectively. However, they also reveal that introducing an auxiliary modality can sometimes degrade performance when dealing with highly informative RGB content, e.g., images acquired in perfectly lit environments. To address this issue, we propose to adapt SAM using \emph{fused} features computed from \emph{both} the RGB and the auxiliary modality, as illustrated by the orange block in \Cref{fig:teaser_network}. This design allows the model to learn to incorporate auxiliary information only when beneficial. For instance, in RGB-LiDAR segmentation, the model can learn to exploit LiDAR information in low-light settings while relying only on RGB features in perfectly lit environments. In summary, our design harnesses the strengths of foundational RGB models and leverages multimodal information to improve performance in challenging scenarios.
We evaluated our proposal on several multimodal benchmarks such as DeLiVER \cite{zhang2023delivering}, FMB \cite{liu2023segmif}, and MUSES \cite{brodermann2024muses}, achieving \textit{state-of-the-art} performance in all the considered datasets. Furthermore, we observe that many existing multimodal benchmarks fail to clearly highlight the benefits of auxiliary sensing modalities, as most data samples can be effectively segmented using only RGB information.
Hence, we also conducted an evaluation using manually crafted test set splits. Specifically, for both DeLiVER and FMB, we divide the test set into two splits: \textit{RGB-easy} and \textit{RGB-hard} (see \Cref{fig:example}). The former includes samples in which the RGB image provides enough information for accurate segmentation. Conversely, in the samples assigned to the latter split, the RGB content alone is poorly informative, making it necessary to rely on the auxiliary modality. By evaluating multimodal approaches on these splits, we better assess their ability to integrate information from multiple sensing modalities synergistically. Notably, our approach consistently performs best on both the \textit{RGB-easy} and \textit{RGB-hard} subsets of the DeLiVER and FMB test sets. 
Furthermore, our multimodal adaptation method, built on the ViT-adapter architecture, outperforms alternative approaches such as LoRA \cite{hu2022lora}, adapted for multimodal input, on the DeLiVER benchmark in the RGB–LiDAR scenario.

To summarize, our contributions are as follows:

\noindent
$\bullet$ We propose to adapt SAM foundational features to pursue multimodal semantic segmentation. In particular, our multimodal adapter, outlined in \Cref{subsec:adapter}, integrates \textit{fused features}, obtained with a fusion module described in \Cref{subsec:Fusion}, with SAM's RGB-only features.


\noindent
$\bullet$ Our approach attains state-of-the-art performance on the DeLiVER \cite{zhang2023delivering}, FMB \cite{liu2023segmif}, and MUSES \cite{brodermann2024muses} benchmarks.

\noindent
$\bullet$ To highlight the effectiveness of multimodal methods in synergistically exploiting the available sensing modalities, we split the DeLiVER and FMB test sets into \textit{RGB-hard} and \textit{RGB-easy} samples and evaluate methods on these splits. Our method performs best in all settings.

\section{Related work}
Our work tackles multimodal semantic segmentation by proposing an adaptation strategy for SAM's foundational features. Hence, in this section, we first review the literature dealing with segmenting images either based on the RGB content or by also deploying auxiliary modalities. Then we pinpoint some recent approaches that attempt to deploy SAM alongside other modalities, as well as previous works aimed at adapting large pre-trained models effectively.
\label{sec:related work}
\subsection{Image Segmentation}
Image segmentation constitutes a fundamental discipline within computer vision, seeking to partition images into coherent and semantically meaningful regions to facilitate detailed analysis. Unlike conventional image classification, which assigns a single label to an entire image, segmentation operates at the pixel level, enabling a comprehensive understanding of visual content and spatial relationships. Segmentation tasks encompass several distinct paradigms such as semantic segmentation, instance segmentation, panoptic segmentation, and salient object detection.
Semantic segmentation assigns a categorical label to each pixel in an image \cite{minaee2021image}. Instance segmentation \cite{hafiz2020survey} distinguishes individual instances of objects belonging to the same class and panoptic segmentation integrates the principles of both semantic and instance segmentation, yielding a unified pixel-wise representation wherein every pixel is attributed either to a specific object instance or a background class. Salient object detection (SOD) \cite{KHAN2024105308} \cite{CHEN2024110600} concentrates on detecting visually prominent objects within a scene, typically generating binary masks that delineate regions most likely to attract human attention. 
In this work, we focus on semantic segmentation, which involves predicting a category label for each pixel. Among segmentation 
During the last years, many semantic segmentation methods have been designed \cite{cheng2021maskformer, cheng2021mask2former, jain2023oneformer, xie2021segformer, chen2023vision}.

Among them, the most relevant to our work is the ViT-adapter \cite{chen2023vision}, which concerns tailoring a Vision Transformer trained for ImageNet classification to handle dense tasks by leveraging a lightweight adapter network.

\begin{figure*}[t]
    \centering
    \includegraphics[width=0.95\linewidth]{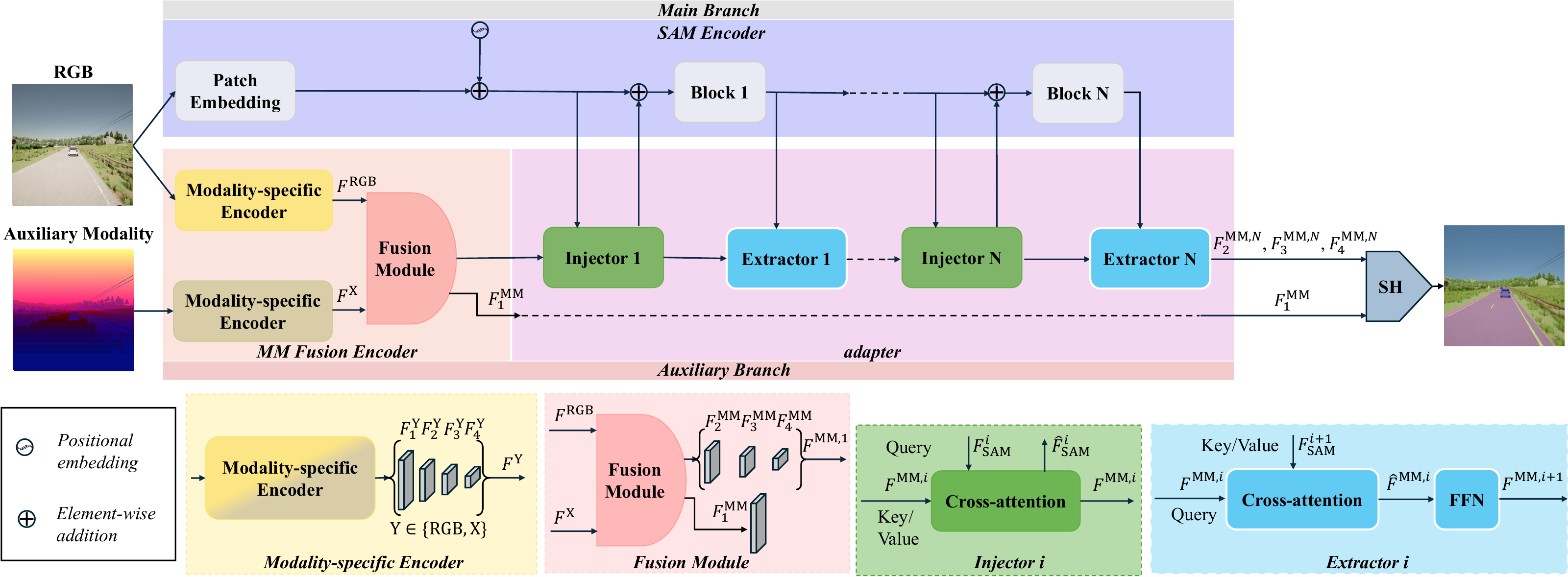}
    \caption{\textbf{Overview of the MM SAM-adapter architecture}. The top row shows the four main modules: SAM Encoder, MM Fusion Encoder, adapter and Segmentation Head (SH). The bottom row details the Modality-specific Encoders and the Fusion Module utilized by the MM Fusion Encoder as well as the Injector and Extractors modules utilized by the adapter.
    }
    \label{fig:MMSAMAD}
\end{figure*}

Thanks to the availability of large annotated RGB datasets, some foundational RGB segmentation models have been recently proposed \cite{Kirillov_2023_ICCV, ravi2024sam, zou2024segment}. 
In particular, the Segment Anything Model (SAM) \cite{Kirillov_2023_ICCV} is designed for prompt-guided instance segmentation, i.e., it can segment objects and their parts in RGB images using prompts such as points, bounding boxes, and masks. Our work aims to rely on SAM's foundational knowledge to pursue semantic segmentation from multiple modalities.

\subsection{Multimodal Semantic Segmentation}
\label{subsec:multimodal_networks}
Multimodal semantic segmentation leverages multiple modalities to address situations where individual sensors may fail \cite{zhang2021deep}. This section focuses on methods that include RGB images as one of the modalities. A key design choice is deciding when to fuse RGB with other modalities.  Early methods \cite{couprie2013indoor} perform channel-wise concatenation of raw input modalities, which are then processed by a single modality-agnostic encoder, while later approaches typically fuse features at multiple levels within the encoder \cite{10.1007/978-3-319-54181-5_14, 8206396, 8666745, zhang2023cmx, zhang2023delivering, pmlr-v235-jia24b, wang2022tokenfusion}. Recent state-of-the-art models  \cite{li2024roadformer, huang2024roadformer+,broedermann2024conditionawaremultimodalfusionrobust} adopt a single late feature fusion stage downstream of the encoder before the decoder. Some methods process the input modalities with a shared modality-agnostic encoder \cite{huang2024roadformer+, li2024stitchfusionweavingvisualmodalities, pmlr-v235-jia24b, broedermann2024conditionawaremultimodalfusionrobust} while others with modality-specific encoders \cite{li2024roadformer, 10.1007/978-3-319-54181-5_14, zhang2023delivering,zhang2023cmx,dong2023efficient}. The machinery to combine features, namely the Fusion Module, has evolved over the years, from simple summation or concatenation to attention mechanisms \cite{zhang2023cmx, pmlr-v235-jia24b}.
Our MM Fusion Encoder deploys modality-specific encoders and late feature fusion. However, peculiar to our design, the purpose of the extracted fused multimodal features is to adjust SAM's RGB features through an adapter network.
The adapter enables the SAM backbone to incorporate auxiliary information at the encoder level, allowing it to leverage multimodal features for semantic segmentation.

\subsection{SAM beyond RGB}
In 2024, several studies extended SAM to tasks involving input modalities beyond RGB \cite{wang2024adaptingsegmentmodelmultimodal, xiao2024segmentmultiplemodalities, Chen_2024_CVPR}.
Chen et al. \cite{Chen_2024_CVPR} distill the SAM backbone into a student network to perform semantic segmentation from event data. Wang et al. \cite{wang2024adaptingsegmentmodelmultimodal} propose a salient object detection framework that uses a fusion module as a prompt generator for the SAM decoder. Xiao et al. \cite{xiao2024segmentmultiplemodalities} introduce a multimodal SAM tailored to prompt-based instance segmentation. 
Yao et al. \cite{10611127} propose a SAM-adapter for the RGB-Event framework, based on cross-attention and gated blocks, where event data are processed through a domain-specific backbone.
Liu et al. \cite{10948332} leverage the full SAM architecture for RGB-Depth and RGB-Event frameworks in the context of mask segmentation. 
Differently, we employ the SAM encoder for multimodal semantic segmentation.
\subsection{Adapters}
\label{subsec:adapters}
Adapters effectively transfer knowledge from powerful backbones to various tasks. Initially introduced in NLP \cite{pmlr-v97-houlsby19a} and later adapted for Computer Vision \cite{xin2024parameterefficientfinetuningpretrainedvision}, they enhance a Transformer's general knowledge by integrating external learnable modules.
Among these methods, LoRA \cite{hu2022lora} is one of the earliest approaches designed for parameter-efficient fine-tuning and has also been extended in the vision domain.
In particular, the ViT-adapter \cite{chen2023vision} aims to adapt Vision Transformers (ViT) pre-trained on image classification for dense prediction tasks. This architecture utilizes a convolutional encoder to extract spatial features, which are integrated with ViT features through a series of injectors and extractors that maintain continuous interaction with the main backbone. Unlike standard adapters, which often freeze the backbone during training, the ViT-adapter fine-tunes the main ViT encoder to achieve optimal performance. Recently, the adapter mechanism has been explored to customize SAM to various downstream tasks \cite{chen2023sam, song2024susamsimpleunifiedframework, chen2024sam2adapterevaluatingadapting, wu2023medicalsamadapteradapting}. However, all previous works focus on adapting SAM to perform tasks that require processing only RGB images. In contrast, we are the first to apply an adapter network to adapt SAM for processing multimodal inputs.
Our adapter mitigates the risk of catastrophic forgetting and enables the integration of multimodal features into the SAM model.

\section{Materials and Method}
\subsection{MM SAM-adapter}
\label{sec:MM SAM-adapter Method}
Our method leverages the benefits of both fusing auxiliary modality features and adaptively integrating knowledge into the SAM backbone. Fusion allows the model to harness auxiliary modality information when the primary modality alone is insufficient for semantic segmentation, while the adaptation technique enables the SAM encoder to incorporate multimodal knowledge, mitigating the risk of catastrophic forgetting in the backbone.  Notably, the overall encoder architecture is asymmetric, as the SAM backbone has a larger number of parameters compared to the combined adapter module and Multimodal Fusion Encoder. The scenario in which this parameter gap is reduced is analyzed in \Cref{subsec:ablation_symm}. Thus, our approach prioritizes the foundational knowledge embedded in the SAM RGB backbone, while exploiting the  \textit{multimodal fused} knowledge captured by the Multimodal Fusion Encoder as well. These advantages enable the model to achieve \textit{state-of-the-art} performance across various benchmarks as discussed in \Cref{sec:experimentalresults} and to perform best in \textit{RGB-easy} and \textit{RGB-hard} scenarios.
As shown in \Cref{fig:MMSAMAD}, our multimodal semantic segmentation framework, MM SAM-adapter, comprises a main branch with the SAM Image Encoder, a secondary branch featuring a Multimodal Fusion Encoder and an adapter module, and a Segmentation Head (SH). The following sections provide a detailed description of these components.
\subsubsection{SAM Encoder}
Our goal is to leverage the segmentation-specific knowledge embedded in SAM's weights. The overall SAM architecture consists of three main modules: an image encoder, a prompt encoder, and a decoder. As the prompt encoder and the decoder are tailored specifically for the prompt-guided instance segmentation task, we deploy only the image encoder in our multimodal semantic segmentation framework. The SAM image encoder is built upon a ViT architecture, which has been pre-trained in a self-supervised manner as a  Masked AutoEncoder \cite{He_2022_CVPR} and then fine-tuned on the SA-1B dataset to perform prompt-guided instance segmentation.  We utilize the SAM encoder based on the ViT-Large (ViT-L) architecture, which includes $L=24$ layers. As presented in \Cref{fig:MMSAMAD} and \Cref{subsec:adapter}, akin to the architecture proposed in \cite{chen2023vision}, we divide the encoder into $N=4$ Blocks consisting of $\frac{L}{N}$ layers, each Block interacting with the adapter by a pair of Injector-Extractor modules.

\subsubsection{Multimodal Fusion Encoder}\label{subsec:Fusion}
As discussed in \Cref{sec:introduction}, our method relies initially on a fusion stage, where RGB features are combined with auxiliary modality features, producing multimodal representations that are subsequently fed into the adapter module.
To achieve this, we employ a Multimodal Fusion Encoder that processes the RGB image and the auxiliary measurements as inputs to generate fused features. Thereby, the Fusion Encoder can learn to weigh the contribution of the different modalities within the signal provided to the adapter, which, in turn, can learn how to integrate it with SAM's features. Our Multimodal Fusion Encoder comprises three main modules: two modality-specific encoders and one fusion module, as described below.

\paragraph{Modality-specific Encoders}\label{subsubsec:Modality-specific encoder}
The modality-specific encoders extract features independently from each of the two modalities to better handle their different nature: e.g., RGB images and LiDAR measurements are dense and sparse signals, respectively.
Inspired by Vit-adapter \cite{chen2023vision}, we employ convolutional networks as they yield spatial features at multiple resolutions that help transformers better capture local spatial information. In particular, we employ ConvNext Small \cite{liu2022convnet} pre-trained on ImageNet-22k \cite{5206848} for both modality-specific encoders. To ensure compatibility, when the auxiliary signal has a single channel, we replicated it three times before feeding it to ConvNext.
Hence, given an RGB image $I^{\text{RGB}}\in \mathbb{R}^{H \times W \times 3}$ and a pixel-aligned auxiliary signal $I^\text{X}\in \mathbb{R}^{H \times W \times 3}$, both encoders produce four different spatial features, namely $F_{i}^{\text{X}}$ and $F_{i}^{\text{RGB}}$ with $i=1,2,3,4$ for the RGB and auxiliary modality encoder, respectively. The feature tensor size at resolution $i$ is  $\frac{H}{2^{i+1}} \times \frac{W}{2^{i+1}} \times D_i$, where  $D_i$ denotes the number of channels (e.g., in ConvNext Small 96, 128, 384, 768).

\paragraph{Fusion Module} \label{subsubsec:Fusion}
The Fusion Module processes multi-scale modality-specific features $(F_i^{\text{X}}, F_i^{\text{RGB}})$ and generates fused multi-scale outputs, $F_{i}^\text{MM}$, for $i=1,2,3,4$. To align with SAM’s feature dimensions, we apply four linear projection layers to match the channel dimension $D$.
The fusion module must preserve information from both modalities to allow the adapter to dynamically select the relevant ones during inference, e.g., only the RGB information in optimal conditions. As shown in \Cref{subsec:ablation_fusion}, to achieve this goal, we can employ a simple feature concatenation of the RGB and auxiliary features, which preserves each modality's information. However, to obtain the best performance, we adopt the Road-Fusion module from Roadformer+ \cite{huang2024roadformer+}, which combines convolutional, self-attention, spatial-attention, and coordinate-attention layers for advanced heterogeneous feature fusion.

\subsubsection{Adapter}
\label{subsec:adapter}
The adapter module takes as input the multimodal features produced by the Multimodal Fusion Encoder and incorporates that knowledge within the SAM backbone. Akin to the ViT-adapter \cite{chen2023vision}, our adapter is a side-tuning network -- a parallel branch that processes information alongside the main backbone. This strategy enables the adapter to learn domain-specific knowledge, in this case, multimodal information, while mitigating the risk of catastrophic forgetting in the main backbone. However, unlike the original proposal in \cite{chen2023vision}, our adapter processes fused multimodal features rather than RGB-only features. 

The adapter consists of a series of two modules: an injector, which introduces spatial multimodal knowledge to the SAM encoder, and an extractor, which retrieves hierarchical features from the SAM backbone.
\begin{figure*}[!h]
    \centering
    \includegraphics[width=0.72\linewidth]{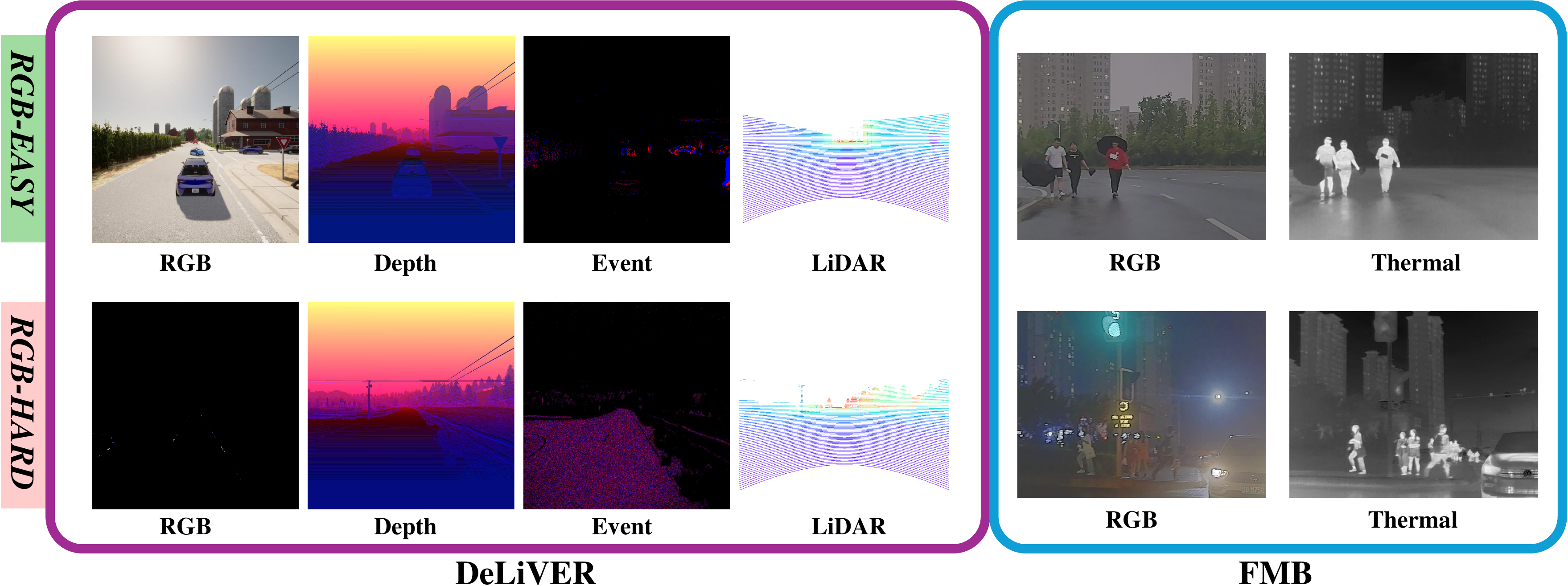}
    \caption{\textit{RGB-easy} and \textit{RGB-hard} examples for DeLiVER and FMB. The DeLiVER \textit{RGB-hard} image has been stretched to enhance visibility (zoom in to better notice it).}
    \label{fig:example}
\end{figure*}

\paragraph{Injector} \label{subsubsec:Injector}
The $F_2^\text{MM}$, $F_3^\text{MM}$, and $F_4^\text{MM}$ multi-scale multimodal fused features are flattened and stacked channel-wise to obtain $F^{MM,1}$, a 2-D array of size $(\frac{HW}{8^2}+\frac{HW}{16^2}+\frac{HW}{32^2}) \times D$. $F^{\text{MM},1}$ will be the input to the first Injector module.
Each injector, \textit{Injector i}, computes a cross-attention between the SAM features, which are output of the block $i - 1$, $F^i_\text{SAM}$ with dimensionality $\frac{HW}{16^2}\times D$, and $F^{\text{MM},i}$, the multiscale multimodal fused features after being processed by $i - 1$ injectors and extractors. In particular, $F^i_\text{SAM}$ is taken as the query and $F^{\text{MM},i}$ is used as the key and value of a multi-scale deformable cross-attention \cite {zhu2021deformable}. Hence, in our adapter, the computation performed by an Injector can be  expressed as:
\begin{equation}
\label{eqn:1}
\centering
\hat{F}^i_\text{SAM}= F^i_\text{SAM} + \gamma_i\text{Attn}(\Phi(F^i_\text{SAM}),\Phi(F^{\text{MM},i}))
\end{equation}
where $\Phi(\cdot)$ denotes the layer normalization operation \cite{ba2016layernormalization}, $\text{Attn}(\cdot)$ represents the multi-scale deformable cross-attention, and $\gamma_i$ is a $D$-dimensional learnable vector initialized to zero to mitigate the impact of the cross-attention at the beginning of the training process. The output from the $i$-th injector  is then processed by the $i$-th SAM block, obtaining $F_\text{SAM}^{i+1}$ with dimensionality $\frac{HW}{16^2}\times D$.

\paragraph{Extractor} \label{subsubsec:Extractor}
This component operates on the output of SAM's blocks. The multi-scale deformable cross-attention \cite{zhu2021deformable} is computed between SAM's features, $F_{\text{SAM}}^{i+1}$, taken as key and values, and the multi-scale fused features $F^{\text{MM},i}$, taken as query. 
The features obtained by the cross-attention, $\hat{F}^{\text{MM},i}$, are fed to a feed-forward network (FFN), obtaining the final features $F^{\text{MM},i+1}$, which will be the input to the next injector module. Hence, the computation performed by an extractor of our adapter can be  expressed as: 

\begin{equation}
\label{eqn:2}
\centering
\hat{F}^{\text{MM},i}= F^{\text{MM},i}+\text{Attn}(\Phi(F^{\text{MM},i}),\Phi(F_\text{SAM}^{i+1}))
\end{equation}
\begin{equation}
\label{eqn:3}
\centering
F^{\text{MM},i+1}=\hat{F}^{\text{MM},i}+FFN(\Phi(\hat{F}^{\text{MM},i}))
\end{equation}
\subsubsection{Segmentation Head}
We obtain the predictions by a Segmentation Head (SH) implemented as a Segformer \cite{xie2021segformer} decoder. It processes the multimodal fused features, $F_1^\text{MM}$, with the adapter tokenized output, $F^{\text{MM},N}$, which is reshaped to obtain feature maps $F^{\text{MM},N}_i, i=2,3,4$ at their original input resolutions.

\subsection{Datasets}
\label{sec:experimental_settings}



We employ the following multimodal semantic segmentation datasets in our experiments:

\noindent
\textbf{DeLiVER \cite{zhang2023delivering}}  is a synthetic dataset comprising 3,983 training, 2,005 validation, and 1,897 test samples. It includes LiDAR, Depth, Event, and RGB data from urban environments with diverse lighting and weather conditions. 
The tensors containing the auxiliary modalities have the same spatial dimensions as the RGB images ($1042\times1042$), with data from all modalities being pixel-wise aligned. 
The number of classes in DeLiVER is 25.

\noindent
\textbf{FMB \cite{liu2023segmif}} contains 1,500 pixel-aligned RGB-Thermal image pairs, 280 of which serve as the test set.  The image pairs are acquired in urban driving scenes with different illumination and weather conditions and have spatial dimensions  800×600.   The number of classes is 14. To select the best checkpoint for all models, we create a validation set by randomly extracting 160 samples from the training set. 

\noindent
\textbf{MUSES \cite{brodermann2024muses}} consists of 1,500 training, 250 validation and 750 test samples collected under various daytime and weather conditions.
It includes LiDAR, Event, Radar, and RGB data, though we excluded Radar due to its sparsity and insufficient information for multimodal segmentation. We align the data from all modalities at pixel level by using the official code \cite{brodermann2024muses}. MUSES employs the same 19 classes as the RGB-only Cityscapes dataset \cite{Cordts2016TheCD}. As the test annotations are withheld, test set evaluations are possible only through online submissions.




\begin{table*}[!h]
\caption{Details of our architecture (SAM Encoder, Modality-Specific Encoders, Fusion Module). $H$ and $W$ represents the Height and Width of the input image, $i \in [1,4]$, $N=4$ and $N_{\text{classes}}$ is the number of classes. \text{SAM}-L encoder has an embedding dimension of 1024.}
    \centering
     \resizebox{0.6755\linewidth}{!}{
    \begin{tabular}{c| c| c| c| c}
        \toprule
        \textbf{Stage} & \textbf{Layer Type} & \textbf{Input} &  \textbf{Output} & \textbf{Output Shape}   \\
        \midrule
         \multicolumn{5}{c}{\cellcolor{SAM}\textbf{SAM Encoder}} \\
        \midrule
        Input  & Image             & -               & $I^\text{\text{RGB}}$       & $H \times W \times 3$ \\
        \hline
       Emb & Patch E., Pos. E.& $I^{\text{\text{RGB}}}$ & $F^{\text{SAM}}_{1}$ & $\frac{HW}{16^2}\times 1024$ \\
        \hline
        Block i &-&  $\hat{F}^{\text{SAM}}_{i}$ & $F^{\text{SAM}}_{i+1}$ & $\frac{HW}{16^2}\times 1024$\\
        \hline
        \midrule
         \multicolumn{5}{c}{\cellcolor{mod}\textbf{Modality-Specific Encoders}} \\
        \midrule
        \multirow{2}{*}{Input}  & Image           & \multirow{2}{*}{-}                           &$I^{\text{\text{RGB}}}$  & $H \times W \times 3$ \\
        & Aux. mod.  &                         &$I^{\text{X}}$  & $H \times W \times 3$ \\
        \hline
        \multirow{4}{*}{\text{\text{RGB}} Enc.}  & \multirow{4}{*}{ConvNeXt }        &\multirow{4}{*}{$I^{\text{RGB}}$}           & $F^{\text{RGB}}_1$ & $ \frac{H}{4}\times \frac{W}{4} \times 96$ \\
        & & & $F^{\text{RGB}}_2$ & $\frac{H}{8} \times \frac{W}{8} \times 192$ \\
         & &  & $F^{\text{RGB}}_3$ &$\frac{H}{16} \times \frac{W}{16} \times 384$ \\
         & &  & $F^{\text{RGB}}_4$ &$\frac{H}{32} \times \frac{W}{32} \times 768$ \\
         \hline
         \multirow{4}{*}{\text{X} Enc.}  & \multirow{4}{*}{ConvNeXt}        &\multirow{4}{*}{$I^{\text{X}}$}             & $F^{\text{X}}_1$& $ \frac{H}{4}\times \frac{W}{4}\times 96$ \\
        & & &$F^{\text{X}}_2$  &$ \frac{H}{8}\times \frac{W}{8} \times 192$ \\
         & & &$F^{\text{X}}_3$ &$ \frac{H}{16}\times \frac{W}{16} \times 384$ \\
         & & &$F^{\text{X}}_4$ &$ \frac{H}{32}\times \frac{W}{32} \times 768$ \\
         \hline
        \midrule
         \multicolumn{5}{c}{\cellcolor{fusion}\textbf{Fusion module}} \\
         \midrule
        \multirow{4}{*}{GFE-\text{RGB}}  & \multirow{2}{*}{Patch E.,}        &  $F^{\text{RGB}}_1$         &$F^{\text{RGB},G}_1$ & $ \frac{H}{4}\times \frac{W}{4} \times 96$ \\
        &  & $F^{\text{RGB}}_2$& $F^{\text{RGB},G}_2$  & $ \frac{H}{8}\times \frac{W}{8} \times 192$ \\
         &MHSA,  & $F^{\text{RGB}}_3$ & $F^{\text{RGB},G}_3$  &$ \frac{H}{16}\times \frac{W}{16} \times 384$ \\
         &LN & $F^{\text{RGB}}_4$ & $F^{\text{RGB},G}_4$ &$ \frac{H}{32}\times \frac{W}{32} \times 768$ \\
         \hline
         \multirow{4}{*}{GFE-X}  & \multirow{2}{*}{Patch E.,}        &$F^{\text{X}}_1$             & $F^{\text{X},G}_1$& $ \frac{H}{4}\times \frac{W}{4} \times 96$ \\
        &   &$F^{\text{X}}_2$ & $F^{\text{X},G}_2$  &$ \frac{H}{8}\times \frac{W}{8} \times 192$ \\
         &MHSA, &$F^{\text{X}}_3$  & $F^{\text{X},G}_3$ &$ \frac{H}{16}\times \frac{W}{16} \times 384$ \\
         &LN &$F^{\text{X}}_4$ & $F^{\text{X},G}_4$ &$ \frac{H}{32}\times \frac{W}{32} \times 768$ \\
         \hline
         \multirow{4}{*}{LFE-\text{RGB}}  & Conv$_{1\times 1}$, RL,         &  $F^{\text{RGB}}_1$         &  $F^{\text{RGB},L}_1$ & $ \frac{H}{4}\times \frac{W}{4} \times 96$ \\
        & DWC$_{3\times 3}$, RL,  & $F^{\text{RGB}}_2$& $F^{\text{RGB},L}_2$ & $ \frac{H}{8}\times \frac{W}{8} \times 192$ \\
         & Conv$_{1\times 1}$ & $F^{\text{RGB}}_3$ & $F^{\text{RGB},L}_3$  &$ \frac{H}{16}\times \frac{W}{16} \times 384$ \\
         & & $F^{\text{RGB}}_4$ & $F^{\text{RGB},L}_4$  &$ \frac{H}{32}\times \frac{W}{32} \times 768$ \\
         \hline
         \multirow{4}{*}{LFE-X}  & Conv$_{1\times 1}$, RL,         &$F^{\text{X}}_1$             & $F^{\text{X},L}_1$& $ \frac{H}{4}\times \frac{W}{4} \times 96$ \\
        & DWC$_{3\times 3}$, RL, &$F^{\text{X}}_2$ &$F^{\text{X},L}_2$  &$ \frac{H}{8}\times \frac{W}{8} \times 192$ \\
         & Conv$_{1\times 1}$&$F^{\text{X}}_3$  &$F^{\text{X},L}_3$ &$ \frac{H}{16}\times \frac{W}{16} \times 384$ \\
         & &$F^{\text{X}}_4$ &$F^{\text{X},L}_4$ &$ \frac{H}{32}\times \frac{W}{32} \times 768$ \\
         \hline
         \multirow{4}{*}{GFRM}  & CA \text{RGB}-X,          & $F^{\text{RGB},G}_1,F^{\text{X},G}_1$                           &$F^{\text{MM},G}_1$  & $ \frac{H}{4}\times \frac{W}{4} \times 192$ \\
        &  CA X-\text{RGB},  & $F^{\text{RGB},G}_2,F^{\text{X},G}_2$                          &$F^{\text{MM},G}_2$  & $ \frac{H}{8}\times \frac{W}{8} \times 384$\\
        & Concat., LN,  & $F^{\text{RGB},G}_3,F^{\text{X},G}_3$                          &$F^{\text{MM},G}_3$  & $ \frac{H}{16}\times \frac{W}{16} \times 768$\\
        & Av. Pool., Sigm.  & $F^{\text{RGB},G}_4,F^{\text{X},G}_4$                          &$F^{\text{MM},G}_4$  & $ \frac{H}{32}\times \frac{W}{32} \times 1536$\\
         \hline
        \multirow{4}{*}{LFFM}  &Concat.,           &  $F^{\text{RGB},L}_1,F^{\text{X},L}_1$                           &$F^{\text{MM},L}_1$  & $\frac{H}{4}\times \frac{W}{4} \times 192$ \\
        & Conv$_{1\times1}$ & $F^{\text{RGB},L}_2,F^{\text{X},L}_2$                         &$F^{\text{MM},L}_2$  & $\frac{H}{8}\times \frac{W}{8} \times 384$\\
        & DWC$_{3\times 3}$, GL, & $F^{\text{RGB},L}_3,F^{\text{X},L}_3$                     &$F^{\text{MM},L}_3$  & $\frac{H}{16}\times \frac{W}{16} \times 768$\\
        & Conv$_{1\times1}$ &$F^{\text{RGB},L}_4, F^{\text{X},L}_4$                      &$F^{\text{MM},L}_4$  & $\frac{H}{32}\times \frac{W}{32} \times 1536$\\
          \hline

          \multirow{4}{*}{FEIM}  &        &  $F^{\text{MM},G}_1,F^{\text{MM},L}_1$                           &$F^{\text{MM},F}_1$  & $\frac{H}{4}\times \frac{W}{4} \times 192$ \\
        &W. Feat. Sum,   & $F^{\text{MM},G}_2,F^{\text{MM},L}_2$                         &$F^{\text{MM},F}_2$  & $\frac{H}{8}\times \frac{W}{8} \times 384$\\
        &Coord. A. & $F^{MM, G}_3,F^{\text{MM},L}_3$                     &$F^{\text{MM},F}_3$  & $\frac{H}{16}\times \frac{W}{16} \times 768$\\
        & &$F^{\text{MM},G}_4, F^{\text{MM},L}_4$                      &$F^{\text{MM},F}_4$  & $\frac{H}{32}\times \frac{W}{32} \times 1536$\\
          \hline
        \bottomrule
    \end{tabular}}
    
\label{tab:nn_architecture}
\end{table*}

\begin{table*}[!h]
\caption{Details of our architecture (Projector, Injector, Extractor, Feature Refinement, Segmentation Head). $H$ and $W$ represent the Height and Width of the input image, $i \in [1,4]$, $N=4$ and $N_{\text{classes}}$ is the number of classes. SAM-L encoder has an embedding dimension of 1024.}
    \centering
     \resizebox{0.784\linewidth}{!}{
    \begin{tabular}{c| c| c| c| c}
        \toprule
        \textbf{Stage} & \textbf{Layer Type} & \textbf{Input} &  \textbf{Output} & \textbf{Output Shape}   \\
          \midrule
         \multicolumn{5}{c}{\textbf{Projector}} \\
        \midrule
             \multirow{4}{*}{S-to-V}  &         &  $F^{\text{MM},F}_1$                           &$F^{\text{MM}}_1$  & $\frac{HW}{4^2}\times 1024$ \\
        & Conv$_{1\times1}$, & $F^{\text{MM},F}_2$                         &$F^{\text{MM}}_2$  & $\frac{HW}{8^2}\times 1024$\\
        & Flatten & $F^{\text{MM},F}_3$                     &$F^{\text{MM}}_3$  & $\frac{HW}{16^2}\times 1024$\\
        & &$F^{\text{MM},F}_4$                      &$F^{\text{MM}}_4$  & $\frac{HW}{32^2} \times 1024$\\
          \hline
          \multirow{3}{*}{Feat. Stack.}  &   \multirow{3}{*}{Concat.}       &  $F^{\text{MM}}_2$                           &\multirow{3}{*}{$F^{\text{MM},1}$}  & \multirow{3}{*}{$(\frac{HW}{8^2}+\frac{HW}{16^2}+\frac{HW}{32^2})\times 1024$}\\
        & & $F^{\text{MM}}_3$                         &  & \\
        & & $F^{\text{MM}}_4$                     &  & \\
          \hline
        \midrule
         \multicolumn{5}{c}{\cellcolor{Inj}\textbf{Injector i}} \\
        \midrule
         
        \multirow{2}{*}{Q-K Inter}     &  MSDA       &  \multirow{2}{*}{$F^{\text{SAM}}_i$,$F^{\text{MM},i}$}                           &\multirow{2}{*}{$\hat{F}^{\text{SAM}}_i$}  & \multirow{2}{*}{$\frac{HW}{16^2}\times 1024$} \\
         & W. Feat. Sum & & \\
          \hline
        \midrule
         \multicolumn{5}{c}{\cellcolor{Ext}\textbf{Extractor i}} \\
         \midrule
         \multirow{2}{*}{Q-K Inter}     &  \multirow{2}{*}{MSDA}         &  \multirow{2}{*}{$F^{\text{MM},i}$,$F^{\text{SAM}}_{i+1}$}                           &\multirow{2}{*}{$\hat{F}^{\text{MM},i}$}  & \multirow{2}{*}{$(\frac{HW}{8^2}+\frac{HW}{16^2}+\frac{HW}{32^2})\times 1024$} \\
         &  & & \\
         \hline
         \multirow{2}{*}{FFN}     &   Lin., DWC$_{3\times3}$      &\multirow{2}{*}{$\hat{F}^{\text{MM},i}$}                  &\multirow{2}{*}{$F^{\text{MM},i+1}$}  & \multirow{2}{*}{$(\frac{HW}{8^2}+\frac{HW}{16^2}+\frac{HW}{32^2})\times 1024$} \\
         & GL, Lin. & & \\
        \midrule
        \midrule
         \multicolumn{5}{c}{\textbf{Feature Refinement}} \\
        \midrule
        \multirow{3}{*}{Slice}  &    \multirow{3}{*}{Reshape}    &  \multirow{3}{*}{$F^{\text{MM},N}$}                          &$F^{\text{MM},N}_2$  & $\frac{H}{8}\times \frac{W}{8} \times 1024$ \\
        &   &                          &$F^{\text{MM},N}_3$  & $\frac{H}{16}\times \frac{W}{16} \times 1024$\\
        & &                      &$F^{\text{MM},N}_4$  & $\frac{H}{32}\times \frac{W}{32} \times 1024$\\
          \hline
         \midrule
          \multicolumn{5}{c}{\cellcolor{seghead}\textbf{Segmentation Head}} \\
        \midrule
        
         \multirow{5}{*}{Prep.}     &   Tran. Conv. \& Sum,      &$F^{\text{MM},N}_2$,$F^{\text{MM}}_1$                  &$F^{\text{MM},N}_1$  & $\frac{H}{4}\times \frac{W}{4}\times 1024$ \\
         & \multirow{4}{*}{Feature Sum} & $F^{\text{MM},N}_1$,$F^{\text{SAM}}_N$ & $F^{\text{MM},N}_{1,\text{mix}}$&$\frac{H}{4}\times \frac{W}{4}\times 1024$ \\
         & & $F^{\text{MM},N}_2$,$F^{\text{SAM}}_N$ &$F^{\text{MM},N}_{2,\text{mix}}$ & $\frac{H}{8}\times \frac{W}{8}\times 1024$ \\
         & & $F^{\text{MM},N}_3$,$F^{\text{SAM}}_N$ &$F^{\text{MM},N}_{3,\text{mix}}$ & $\frac{H}{16}\times \frac{W}{16}\times 1024$  \\
         & & $F^{\text{MM},N}_4$,$F^{\text{SAM}}_N$ &$F^{\text{MM},N}_{4,\text{mix}}$ & $\frac{H}{32}\times \frac{W}{32}\times 1024$ \\
         \midrule
         \multirow{4}{*}{\text{up}}     &   Conv$_{1\times 1}$,      &$F^{\text{MM},N}_{1,\text{mix}}$                  &\multirow{4}{*}{$F^{\text{MM}}_{\text{up}}$}  & \multirow{4}{*}{$\frac{H}{4}\times \frac{W}{4}\times 512$} \\
         & BN, RL, & $F^{\text{MM},N}_{2,\text{mix}}$& \\
         & Interp., & $F^{\text{MM},N}_{3,\text{mix}}$& \\
         & Concat.& $F^{\text{MM},N}_{4,\text{mix}}$ & \\
         \hline
         \multirow{4}{*}{Pred.}     &   Conv$_{1\times 1}$,      &\multirow{4}{*}{$F^{\text{MM}}_{\text{up}}$}                 &\multirow{4}{*}{$F^{\text{MM}}_{\text{Seg}}$}  & \multirow{4}{*}{$H\times W\times N_{\text{classes}}$} \\
         & BN, RL, & & \\
         & Conv$_{1\times 1}$, & & \\
         & Interp.&  & \\
        \bottomrule
    \end{tabular}}
    
\label{tab:nn_architecture_part2}
\end{table*}
\begin{table}[h!]
\caption{Legend of abbreviations used in the \Cref{tab:nn_architecture} and \Cref{tab:nn_architecture_part2}} 
    \centering
   \resizebox{\linewidth}{!}{
    \begin{tabular}{c|l}
        \hline
        \textbf{Abbreviation} & \textbf{Meaning} \\ 
        \hline
        Emb & Embedding\\
        \hline
        Patch E. & Patch Embedding\\
        \hline
        Pos. E. & Positional Embedding\\
        \hline
        Aux mod & Auxiliary Modality\\
        \hline
        RGB Enc. & RGB Encoder\\
        \hline
        X Enc. & X Encoder\\
        \hline
        GFE  & Global Feature Enhancer \\ 
        \hline
        MHSA & Multi-Head Self-Attention\\
        \hline
        LN & Layer normalization\\
        \hline
        Concat. & Concatenation\\
        \hline
        LFE  & Local Feature Enhancer \\ 
        \hline
        DWC$_{3\times3}$ & Depth-wise Separable Convolution \\
        \hline
        Conv$_{1\times1}$ & $1\times1$ Convolution \\
        \hline
        GFRM & Global Feature Recalibration Module \\
        \hline
        Sigm. & Sigmoid\\
        \hline
        Av. Pool. & Average Pooling \\
        \hline
        CA  & Cross-Attention \\ 
        \hline
        LFFM & Local Feature Fusion Module\\
        \hline
        FEIM & Feature Enhancement and Integration Module\\
        \hline
        W. Feat. Sum & Weigthed Feature Sum\\
        \hline
        Coord. A. & Coordinate Attention\\
        \hline
        MSDA  & Multi-Scale Deformable Attention\\ 
        \hline
        BN  & Batch Normalization \\ 
        \hline
        Prep. & Preprocessing\\
        \hline
        Up & Upsampling\\
        \hline
        Pred. & Prediction\\
        \hline
        Tran. Conv. & Transpose Convolution\\
        \hline
        Q-K Inter & Query-Key Interaction\\
        \hline
        S-to-V & Spatial-to-Vector\\
        \hline
        Feat. Stack. & Feature Stacking\\
        \hline
        Interp. & Interpolation\\
        \hline
        Lin. & Linear\\
        \hline
        RL &ReLU\\
        \hline
        GL &GeLU\\
        \hline
    \end{tabular}}
     
     \label{tab:legend}
\end{table}

\begin{table*}[t]
\caption{\textbf{DeLiVER test set results} in the RGB-Depth (\textcolor{darkBlue}{\textbf{RGB-D}}), RGB-LiDAR (\textcolor{darkRed}{\textbf{RGB-L}}), and RGB-Event (\textcolor{darkGreen}{\textbf{RGB-E}}) setups.}
    \centering
    \resizebox{0.85\linewidth}{!}{
    \begin{tabular}{l|ccc|ccc|ccc}
        \toprule
        \multirow{2}{*}{\textbf{Method}} & \multicolumn{3}{c|}{\cellcolor{cyan20}\textbf{All}} & \multicolumn{3}{c|}{\cellcolor{PeaGreen}\textbf{RGB-easy}} & \multicolumn{3}{c}{\cellcolor{red20}\textbf{RGB-hard}} \\
         & \textcolor{darkBlue}{\textbf{RGB-D}} & \textcolor{darkRed}{\textbf{RGB-L}} & \textcolor{darkGreen}{\textbf{RGB-E}} & 
                         \textcolor{darkBlue}{\textbf{RGB-D}} & \textcolor{darkRed}{\textbf{RGB-L}} & \textcolor{darkGreen}{\textbf{RGB-E}} & 
                         \textcolor{darkBlue}{\textbf{RGB-D}} & \textcolor{darkRed}{\textbf{RGB-L}} & \textcolor{darkGreen}{\textbf{RGB-E}} \\
        \midrule
        CMNeXt \cite{zhang2023delivering} 
        & 53.87 & 51.32 & 50.81 & 54.07 & 51.97 & 51.56 & 49.66 & 39.29 & 37.08 \\
        GeminiFusion \cite{pmlr-v235-jia24b} 
        & 54.98 & 50.57 & 51.71 & 55.07 & 51.07 & 52.32 & 53.00 & 40.93 & 40.27 \\
        RoadFormer+ \cite{huang2024roadformer+}
        & 55.95 & 54.56 & 54.10 & 56.12 & 55.41 & 54.80 & 52.39 & 40.26 & 41.98 \\
        \textbf{MM SAM-adapter} & \textbf{57.35} & \textbf{57.14} & \textbf{55.70} & \textbf{57.62} & \textbf{57.75} & \textbf{56.29} & \textbf{53.35} & \textbf{45.46} & \textbf{44.67} \\
        \bottomrule
    \end{tabular}}
    
    \label{tab:DeLiVER}
\end{table*}

\subsubsection{RGB-easy and RGB-hard splits} \label{subsubsec:easy_hard}
As highlighted in \Cref{sec:introduction}, most samples in multimodal datasets are easily segmented by leveraging exclusively  RGB information. For instance, from our experiments, approximately 97\% of DeLiVER sample can be optimally segmented using solely the RGB image, while only the remaining 3\% represents challenging scenarios in which the auxiliary information may be useful.
To better assess the methods' capability to exploit information from multiple modalities synergistically, we divided the DeliVER and FMB test sets into an \textit{RGB-hard} and \textit{RGB-easy} split. Purposely, we first train an RGB-only network on each multimodal dataset to obtain semantic predictions. Then, we visually compare the RGB-only predictions to the ground-truths while paying attention to the information conveyed by the auxiliary modality. Thereby, we identify as \textit{RGB-hard} those samples in which the RGB-only network fails to recognize elements that may be correctly detected by another modality, and as \textit{RGB-easy} all other samples. 
%
Accordingly, the  DeLiVER test set is divided into 1797 \textit{RGB-easy} and 100 \textit{RGB-hard} samples, whereas for FMB we get  183 and 97 samples for the \textit{RGB-easy} and \textit{RGB-hard} splits. 
%
\Cref{fig:example} shows samples for the \textit{RGB-easy} and \textit{RGB-hard} splits of DeLiVER and FMB. We point out that we adopt a similar procedure to build a balanced validation set of FMB, i.e., we randomly select 80 \textit{RGB-easy} and 80 \textit{RGB-hard} samples from the original training set. 
We did not perform a similar test set split on MUSES \cite{brodermann2024muses} as the ground truths are unavailable.
Yet, we report results for all the official splits (day, night, different weather conditions). In particular,  this disentangled evaluation allows us to assess the behaviour of the methods in the fog and nighttime scenarios -- those that pose more challenges for the RGB modality, as highlighted in the original paper \cite{brodermann2024muses}.

\subsection{Implementation Details}\label{sec:Implementation Details}


\subsubsection{Architectural Details}\label{sec:Architectural Details}
The architectural specifications are summarized in \Cref{tab:nn_architecture} \hl{and} \Cref{tab:nn_architecture_part2}, which offers a detailed overview of our network design. A legend explaining the abbreviations used in \Cref{tab:nn_architecture} \hl{and} \Cref{tab:nn_architecture_part2} is provided in \Cref{tab:legend}.

The \textbf{Fusion Module}, as detailed in \Cref{tab:nn_architecture} and employed throughout this paper,
is based on the RoadFormer+ fusion block \cite{huang2024roadformer+} (Road-fusion module). This module comprises several key components, including two Global Feature Enhancers (GFEs), two Local Feature Enhancers (LFEs), one Global Feature Recalibration Module (GFRM), one Local Feature Fusion Module (LFFM) and one Feature Enhancement and Integration Module (FEIM).  The Road-fusion block initially extracts global and local features separately from RGB features and auxiliary modality features, both originating from the two Modality-Specific Encoders, by employing GFEs and LFEs. The former are a transformer-based modules, while the latter are convolution-based. Subsequently, the global RGB features and auxiliary modality features are fused using the GFRM. This module performs cross-attention operations between the two modalities: one operation utilizes RGB modality features as queries and auxiliary modality features as keys/values, whereas the other operation reverses these roles. The resulting global feature representations are then concatenated, and the obtained multimodal representation is further refined through an average pooling layer followed by a sigmoid activation function. In parallel with the global features extraction, the local features from both modalities are fused using the LFFM. This module first performs a channel-wise concatenation, after which the resulting feature map is processed through a series of convolutional layers employing one GeLU activation function. Subsequently, local and global fused features are combined through the GFRM. This module employs coordinate attention to effectively encode spatial relationships within the features, enhancing their representational capacity. The resulting attention-enhanced representation is further refined through a series of convolutional layers. The output of the fusion module is subsequently passed through a series of projection layers to align the channel dimensions of the fused features with those of SAM.

The \textbf{Segmentation Head} (SH) will take as input $F_1^\text{MM}$ and the adapter output, $F^{\text{MM},N}$ reshaped into $F^{\text{MM},N}_i, i=2,3,4$.
The $F_1^\text{MM}$ is summed with $F_2^{\text{MM},N}$, which has been upsampled using transpose convolution \cite{zeiler2010deconvolutional} to match the spatial resolution, resulting in $F_1^{\text{MM},N}$. Each spatial feature is subsequently enriched by summing it with the SAM tokens from the final layer, which are first interpolated to match the spatial resolution. This results in the mixed feature representation $F^{\text{MM},N}_{i,\text{mix}}$, where $i = 2, 3, 4$.
Subsequently, the Segformer head \cite{xie2021segformer} processes each spatial feature $F^{\text{MM},N}_{i,\text{mix}}$, using distinct MLPs, composed of $1\times 1$ convolutional layers, batch normalization layers and ReLU activation functions. A channel-wise concatenation operation with bilinear interpolation is then applied, yielding $F^\text{MM}_{\text{up}}$ with dimensionality $\frac{H}{4}\times \frac{W}{4} \times 512$. Finally, in the last stage, another MLP is used to generate the segmentation prediction, with dimensionality $H\times W \times N_{\text{classes}}$, where $N_{\text{classes}}$ denotes the number of classes. 
\subsubsection{Training Details}\label{sec:Training Details}
Our method was trained for 100 epochs using the OHEM cross-entropy loss on two NVIDIA RTX 3090 GPUs with batch size 8 -- we employ gradient accumulation for memory constraints --, \hl{a base learning rate of $2e^{-4}$, which follows a polynomial strategy with a power of 0.9, and exponential warm-up with base 0.1. Therefore, the learning rate evolves according to the following formula:}
\begin{equation}
\eta(p) =
\begin{cases}
\eta_{\text{base}} \cdot 
\left(\text{wr}\right)^{\,1 - \frac{p}{N_{\text{w}}}},
& \text{if } p \le N_{\text{w}}, \\[10pt]
\left(\eta_{\text{base}} - \eta_{\text{min}}\right)
\cdot \left(1 - \frac{p}{P_{\max}}\right)^{\alpha}
+ \eta_{\text{min}}, 
& \text{if } p > N_{\text{w}}.
\end{cases}
\end{equation}
\hl{In the learning rate schedule described above, $\eta(p)$ denotes the learning rate at epoch or iteration $p$. The parameter $\eta_{\text{base}}$ represents the base learning rate that the schedule aims to reach at the end of the warm-up phase ($2e^{-4}$), while $\eta_{\text{min}}$ defines the minimum learning rate that serves as a lower bound throughout training ($\eta_{\text{min}}=0$). During the warm-up phase, which lasts for $N_{\text{w}}$ epochs ($N_{\text{w}}=10$), the learning rate increases exponentially from an initial value determined by scaling $\eta_{\text{base}}$ by the factor $\text{wr}$, which represent the warm-up ratio ($\text{wr}=0.1$). Specifically, at the beginning of training, the learning rate is set to $\eta_{\text{base}} \times \text{wr}$, and it grows smoothly and exponentially toward $\eta_{\text{base}}$ as training progresses through the warm-up steps. After the warm-up period, the learning rate follows a polynomial decay controlled by the exponent $\alpha$ ($\alpha=0.9$), which determines how sharply the learning rate decreases. The variable $P_{\max}$ specifies the total number of epochs in the training process, ensuring that the schedule spans the entire duration of learning without abrupt transitions.}

The optimizer used during training is AdamW \cite{loshchilov2018decoupled} with a weight decay of $1e^{-2}$. 
We use a layer-wise learning rate decay of 0.9, as in ViT-adapter architecture\cite{chen2023vision}. 
 \hl{In this approach, the learning rate assigned to each layer decreases exponentially as one moves from the higher layers toward the lower layers of the backbone. The learning rate for the parameters in layer $\ell$ is defined as:}

\begin{equation}
\eta_\ell = \eta_{\text{curr}} \times \gamma^{L - \ell - 1},
\end{equation}

\hl{where $\eta_{\text{curr}}$ denotes the learning rate assigned to the highest layer, $\gamma$ is the decay rate ($\gamma=0.9$), $L$ is the total number of layers of the transformer backbone, and $\ell$ is the index of the current layer, starting from zero. 

Parameters associated with embeddings, special tokens, and normalization layers are assigned to layer zero and are typically exempted from weight decay. Conversely, the newly introduced layers in the architecture, such as adapter modules and decoding heads, are trained with higher learning rates to encourage rapid adaptation to the downstream task. This parameter grouping ensures that lower layers of the backbone, which encode generic visual features learned during pre-training, are updated more conservatively, thereby preserving their valuable representations.}
In order to improve robustness of our model, we perform online data augmentations based on random resize with a ratio from 0.5 to 2.0, random horizontal flipping, photometric distortion, random Gaussian blur with probability $p$=0.2, and random crop to $1024\times1024$ resolution for DeLiVER and MUSES,  $800\times800$ for FMB. \hl{Our method on FMB has been trained for 200 epochs.}\\
\subsubsection{Competitor Details}
For competitors evaluation, we use official weights when available; otherwise, we train them following the official paper guidelines. 
In the \textbf{DeLiVER} benchmark, CMNeXt and GeminiFusion were evaluated using the official weights, which are based on the most effective backbone, MiT-B2 \cite{xie2021segformer}. In contrast, we trained RoadFormer+ from scratch, as no pre-trained weights were available, using ConvNeXt-Large, the best-performing configuration reported in the paper.
For the \textbf{FMB} \cite{liu2023segmif} and \textbf{MUSES} \cite{brodermann2024muses} datasets, we had to train all competitors due to the unavailability of pre-trained weights. Specifically, for \textbf{FMB}, we followed the official guidelines to ensure optimal performance, training CMNeXt with the MiT-B5 \cite{xie2021segformer} backbone, GeminiFusion with Swin Transformer Large \cite{liu2021Swin}, and RoadFormer+ with ConvNeXt-Large \cite{liu2022convnet}.
Regarding \textbf{MUSES}, since it was recently published with no available training guidelines, we adopted the same configuration as \textbf{DeLiVER}, given the similarity between the datasets in terms of modalities and high-resolution images. Consequently, we trained CMNeXt and GeminiFusion with the MiT-B2 backbone and RoadFormer+ with the ConvNeXt-Large backbone. 
CAFuser \cite{broedermann2024conditionawaremultimodalfusionrobust}, according to its authors, is inherently a multimodal model. Moreover, its training requires an additional text prompt describing the condition of the scene, which is not provided in \textbf{FMB}, but can be implicitly inferred from \textbf{DeLiVER} and \textbf{MUSES}.
Thus, the results are reported following the original paper’s settings.



We highlight that, differently from most competitors that employ different backbones depending on the dataset to achieve the best performance, such as CMNeXt \cite{zhang2023delivering} and GeminiFusion \cite{pmlr-v235-jia24b}, our framework remains consistent across datasets yet achieves superior performance.

\section{Results}\label{sec:experimentalresults}
This section compares MM SAM-adapter with \textit{state-of-the-art} multimodal semantic segmentation models such as CMNext \cite{zhang2023delivering}, GeminiFusion \cite{pmlr-v235-jia24b}, RoadFormer+ \cite{huang2024roadformer+}, and CAFuser \cite{broedermann2024conditionawaremultimodalfusionrobust}. 
We use Mean Intersection Over Union (mIoU) as the semantic segmentation evaluation metric.
In addition, we analyze the design choices underlying the proposed multimodal adapter.

\begin{table}[t]
\caption{\textbf{FMB test set results} in the RGB-Thermal (\textcolor{YellowOrange}{\textbf{RGB-T}}) setup across different scenarios.}
    \centering
    \resizebox{0.85\linewidth}{!}{
    \begin{tabular}{l|c|c|c}
        \toprule
        \multirow{2}{*}{\textbf{Method}} 
        & \cellcolor{cyan20}\textbf{All} & \cellcolor{PeaGreen}\textbf{RGB-easy} & \cellcolor{red20}\textbf{RGB-hard} \\
         & \textcolor{YellowOrange}{\textbf{RGB-T}} & \textcolor{YellowOrange}{\textbf{RGB-T}} & \textcolor{YellowOrange}{\textbf{RGB-T}} \\
        \midrule
        CMNeXt \cite{zhang2023delivering}& 61.66 & 64.81 & 56.85 \\
        GeminiFusion \cite{pmlr-v235-jia24b}& 64.75 & 68.05 & 61.03 \\
        RoadFormer+ \cite{huang2024roadformer+}& 64.57 & 66.31 & 61.79 \\
        \textbf{MM SAM-adapter} & \textbf{66.10} & \textbf{68.45} & \textbf{62.59} \\
        \bottomrule
    \end{tabular}}
    
    \label{tab:FMB}
\end{table}
\begin{table*}[t]
\caption{\textbf{MUSES test results} in the RGB-LiDAR (\textcolor{darkRed}{\textbf{RGB-L}}) and RGB-Event (\textcolor{darkGreen}{\textbf{RGB-E}}) setups for different weather conditions.}
    \centering
    \resizebox{\linewidth}{!}{ 
    \begin{tabular}{l|cc|cc|cc|cc|cc|cc|cc}
        \toprule
        \multirow{2}{*}{\textbf{Method}} & \multicolumn{2}{c|}{\textbf{All}} & \multicolumn{2}{c|}{\textbf{Day}} & \multicolumn{2}{c|}{\textbf{Night}} & \multicolumn{2}{c|}{\textbf{Clear}} & \multicolumn{2}{c|}{\textbf{Fog}} & \multicolumn{2}{c|}{\textbf{Rain}} & \multicolumn{2}{c}{\textbf{Snow}} \\
        & \textcolor{darkRed}{\textbf{RGB-L}} &  \textcolor{darkGreen}{\textbf{RGB-E}} & \textcolor{darkRed}{\textbf{RGB-L}} &  \textcolor{darkGreen}{\textbf{RGB-E}} & \textcolor{darkRed}{\textbf{RGB-L}} &  \textcolor{darkGreen}{\textbf{RGB-E}} & \textcolor{darkRed}{\textbf{RGB-L}} &  \textcolor{darkGreen}{\textbf{RGB-E}} & \textcolor{darkRed}{\textbf{RGB-L}} &  \textcolor{darkGreen}{\textbf{RGB-E}} & \textcolor{darkRed}{\textbf{RGB-L}} &  \textcolor{darkGreen}{\textbf{RGB-E}} & \textcolor{darkRed}{\textbf{RGB-L}} &  \textcolor{darkGreen}{\textbf{RGB-E}} \\
        \midrule
        CMNeXt \cite{zhang2023delivering} & 72.36 & 70.49 & 75.42 & 74.02 & 64.27 & 60.72 & 72.81 & 71.24 & 59.98 & 60.91 & 72.48 & 68.78 & 71.92 & 70.54 \\
        GeminiFusion \cite{pmlr-v235-jia24b}& 74.22 & 68.62 & 75.90 & 72.04 & 67.67 & 58.19 & 74.79 & 70.40 & 62.95 & 60.53 & 73.30 & 65.27 & 74.16 & 68.87 \\
        RoadFormer+ \cite{huang2024roadformer+} & 80.38 & 77.70 & 82.56 & 80.62 & 74.51 & 69.02 & 79.39 & 78.44 & 69.64 & \textbf{69.29} & \textbf{80.69} & 76.40 & 79.49 & 77.70 \\
        \textbf{MM SAM-adapter} & \textbf{81.07} & \textbf{79.92} & \textbf{83.34} & \textbf{83.39} & \textbf{74.97} & \textbf{72.38} & \textbf{80.82} & \textbf{81.52} & \textbf{74.12} & 68.97 & 80.00 & \textbf{78.92} & \textbf{80.85} & \textbf{78.72} \\
        \bottomrule
    \end{tabular}}
    
    \label{tab:MUSES}
\end{table*}
\begin{table}[t]
\caption{\textbf{Comparison with all-modalities competitors on the DELIVER test (Test mIoU) and validation (Val mIoU) sets.}} 
\centering
    \resizebox{\linewidth}{!}{
    \begin{tabular}{l|c|c|c}
        \toprule
         \textbf{Method}  & \textbf{Modalities} & \textbf{Val mIoU} & \textbf{Test mIoU} \\
         \midrule
        CMNeXt \cite{zhang2023delivering}    &RGB, Depth, LiDAR, Event & 66.30 & 53.00\\
        GeminiFusion \cite{pmlr-v235-jia24b}   &RGB, Depth, LiDAR, Event & 66.90 & 54.46 \\
        CAFuser \cite{broedermann2024conditionawaremultimodalfusionrobust} &RGB, Depth, LiDAR, Event &67.80 & 55.60 \\
        \midrule
        MM SAM-adapter   & RGB, LiDAR & 61.89 &  57.14  \\
        MM SAM-adapter   & RGB, Event & 60.74 &  55.70  \\
        \textbf{MM SAM-adapter}   & RGB, Depth & \textbf{69.60} &  \textbf{57.35}  \\
         \bottomrule 
    \end{tabular}}
    
    \label{tab:DELIVER_all}
\end{table}
\begin{table}[t]
    \caption{\textbf{Comparison with all-modalities competitors on the MUSES test set.}}
    \centering
    \resizebox{0.85\linewidth}{!}{
    \begin{tabular}{l|c|c}
        \hline
         \textbf{Method}  & \textbf{Modalities} & \textbf{mIoU} \\\hline
        CMNeXt 
        & RGB, LiDAR, Event, Radar & 72.40\\
        GeminiFusion 
        & RGB, LiDAR, Event, Radar & 75.30\\
        CAFuser 
        & RGB, LiDAR, Event, Radar & 78.18 \\
        \hline
        \textbf{MM SAM-adapter}   & RGB, Event &  \textbf{79.92}  \\
        \textbf{MM SAM-adapter}   & RGB, LiDAR &  \textbf{81.07}  \\
         \hline 
    \end{tabular}}

    \label{tab:MUSES_all}
\end{table}

\subsection{Main results}\label{subsec:Main_results}
\Cref{tab:DeLiVER} reports the results on the DeLiVER test set in three multimodal setups: RGB-Depth (RGB-D), RGB-LiDAR (RGB-L), and RGB-Events (RGB-E). 
By analyzing the performance in the \textit{RGB-hard} case, we note that methods are more effective when employing the Depth auxiliary modality rather than Event or LiDAR data. We ascribe it to the perfect synthetic Depth maps provided with  DeLiVER, which convey extremely rich auxiliary information. In this RGB-D scenario, the improvements over competitors are marginal (e.g., 53.35 vs. 53.00 mIoU of our method vs. GeminiFusion). However, our method shines when the auxiliary modality is more noisy and realistic, such as in the RGB-L and RGB-E setups. In these cases, we achieve large performance improvements over competitors (e.g., 45.46 vs. 40.93 mIoU of our method vs. GeminiFusion in the RGB-L scenario).
Finally, by examining our method's results in the \textit{RGB-easy} scenario across the RGB-D, RGB-L, and RGB-E setups, we note that it achieves very similar performance independently of the auxiliary modality (i.e., 57.62, 57.75, 56.29 mIoU in the RGB-D, RGB-L, and RGB-E setups, respectively). Conversely, other methods generally perform less consistently (e.g., CMNeXt obtains 54.07, 51.97, and 51.56 in the RGB-D, RGB-L, and RGB-E setups, respectively). In general, these results highlight how our method more effectively utilizes the synergies between modalities,   e.g., in the case of good RGB images, the auxiliary modality contribute less to final results. \hl{Our method excels at fusing multimodal features and seamlessly adapting the SAM RGB backbone to incorporate this information.}
\hl{As can be seen from these results, our method establishes a new \textit{state-of-the-art} on the DeLiVER test set in the RGB-D, RGB-L, and RGB-E setups, and achieves remarkable performance in both \textit{RGB-easy} and \textit{RGB-hard} scenarios.}

\Cref{tab:FMB} shows the results on the FMB test set, which features RGB and Thermal images (RGB-T). Remarkably, our method achieves \textit{state-of-the-art} performance also in this real dataset. We note a significant performance drop between \textit{RGB-easy} and \textit{RGB-hard} samples, which validates our manually defined splits. \hl{As evidenced by the results, in the \textit{RGB-hard} scenario, our method demonstrates superior exploitation of auxiliary modality information, yielding a measurable performance gain (e.g., 62.59 vs. 61.79 mIoU of our method vs RoadFormer+). In the \textit{RGB-easy} scenario, our approach likewise surpasses competing methods, though with a narrower margin (e.g., 68.45 vs. 68.05 mIoU of our method vs. GeminiFusion).}

\Cref{tab:MUSES} presents the result on the MUSES dataset \cite{brodermann2024muses} for the RGB-LiDAR (RGB-L) and RGB-Event (RGB-E) setups. We highlight that the results of each model were submitted to the MUSES online benchmark to construct this table. Notably, we achieve \textit{state-of-the-art} performance overall and in most daytime and weather conditions. In particular, we highlight how, in the foggy scenario, one of the most challenging for the RGB modality alone, our method largely surpasses the best competitor performance in the RGB-L setup (i.e., 74.12 vs 69.64 mIoU of our method vs RoadFormer+) and provides by far the most effective solution to tackle this difficult setting. \hl{In the nighttime scenario—also a challenging condition—our method consistently outperforms prior approaches, achieving higher mIoU scores (e.g., 72.48 vs. 69.02 mIoU in the RGB-E setting and 74.97 vs. 74.51 mIoU in the RGB-L setting, of our method vs. RoadFormer+). Remarkably, our method demonstrates strong performance under daytime conditions, a scenario where RGB images are particularly informative (e.g., 83.39 vs. 80.62 mIoU in the RGB-E Day setting, and 83.34 vs. 82.56 mIoU in the RGB-L Day setting, of our method vs RoadFormer+).}

\begin{figure*}[h!]
    \centering
    \includegraphics[width=0.70\linewidth]{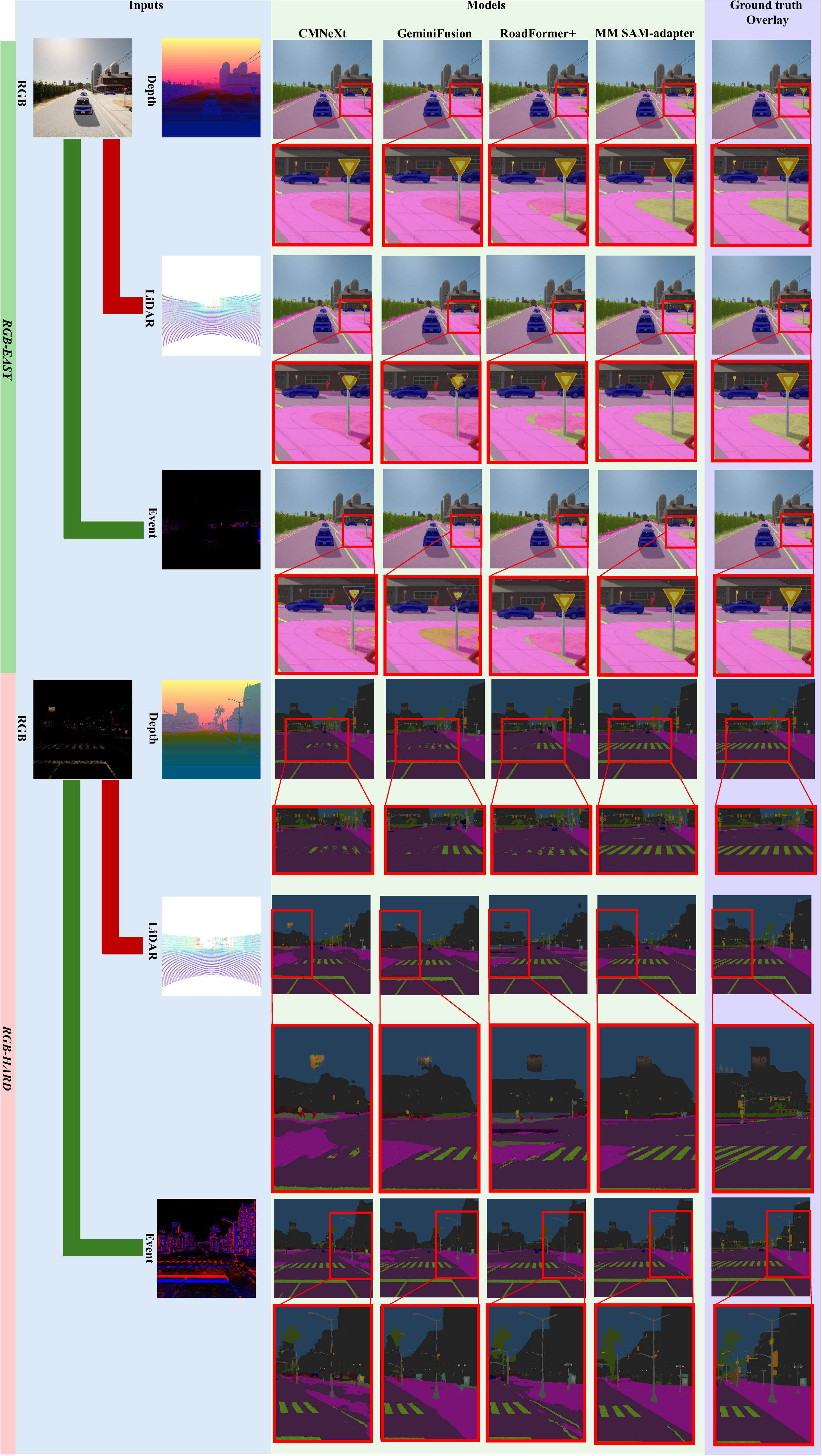}
    \caption{\textbf{DeLiVER \cite{zhang2023delivering} test set predictions} in RGB-Depth, RGB-LiDAR, RGB-Event framework. \textit{RGB-easy}  (top) and \textit{RGB-hard} (bottom) samples. Notably, the input RGB image in \textit{RGB-hard} case has been stretched with an exponential operator.}
    \label{fig:qualitatives_DEPTH}
\end{figure*}

\begin{figure*}[!h]
    \centering
\includegraphics[width=0.77\linewidth]{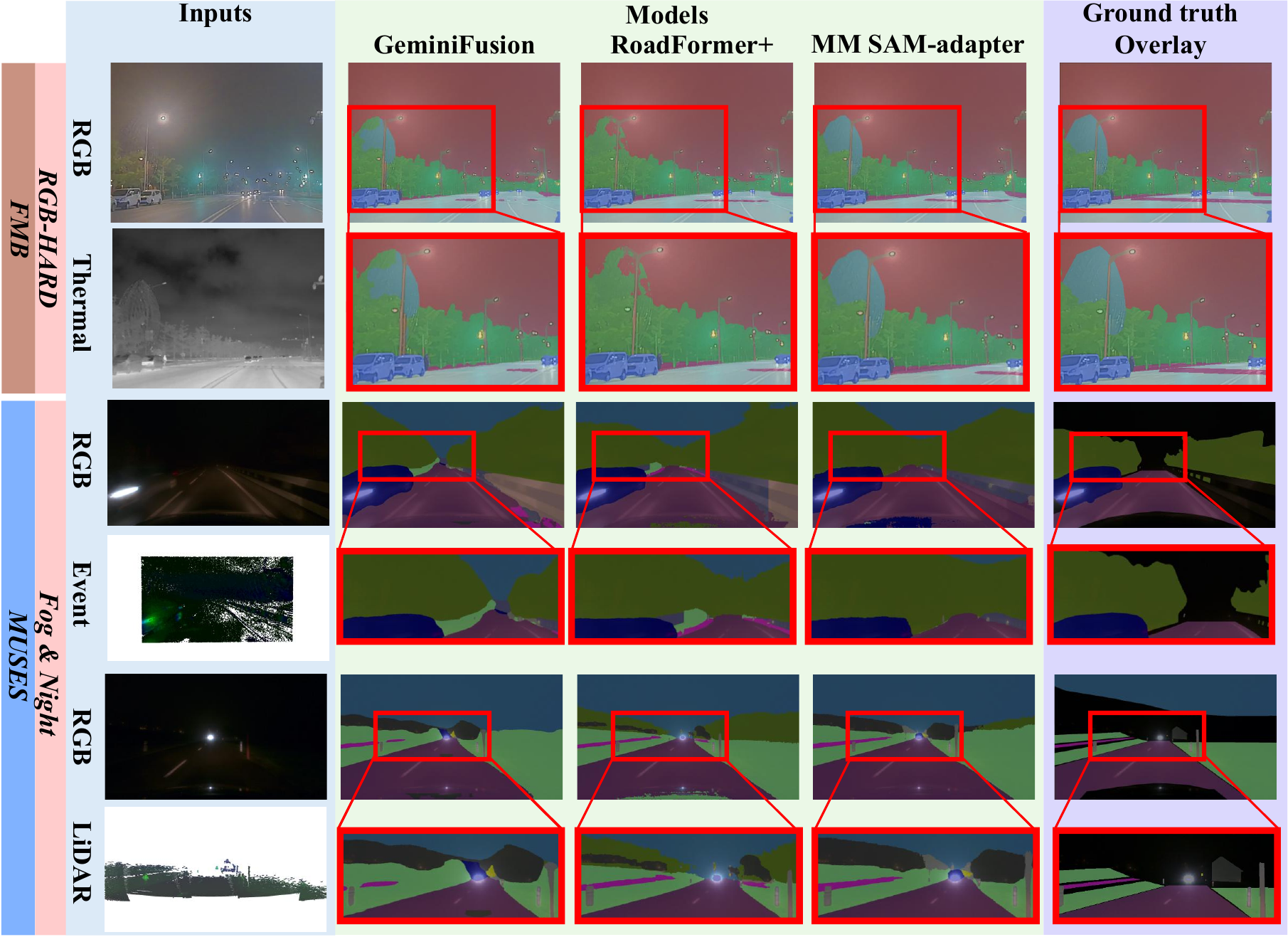}
    \caption{\textbf{FMB test set and MUSES validation set predictions of the methods in \Cref{tab:FMB} and in \Cref{tab:MUSES}}. Black pixels in the last column denote missing ground truth labels.}
    \label{fig:qualitatives_real}
\end{figure*}

\hl{Qualitative results of our method and the competing models are shown in} \Cref{fig:qualitatives_DEPTH} 
, in \Cref{fig:qualitatives_real}.
A closer inspection of the qualitative results in \Cref{fig:qualitatives_DEPTH}
reveals that our method demonstrates a superior ability to capture fine-grained scene details, including road features (e.g., zebra crossings and lane markings), sidewalk elements (e.g., streetlights), and subtle distinctions between contiguous surfaces (e.g., terrain and concrete). Notably, our method produces more accurate qualitative predictions than competing approaches in both the \textit{RGB-hard} and \textit{RGB-easy} settings. In \Cref{fig:qualitatives_real}, which illustrates examples from both FMB and MUSES under the \textit{RGB-hard} condition, our model exhibits a stronger understanding of the overall environment and background elements. In particular, the building partially occluded by trees in the FMB \textit{RGB-hard} scenario is segmented with noticeably higher precision compared to the competing methods. 
\hl{These qualitative results highlight the robustness of our method in various scenarios and datasets, aligning with the findings from the quantitative results.}
Additional examples are provided in \Cref{appendix}.

\subsection{All-modalities competitors}\label{subsec:All_modalities_competitors}
This section compares MM SAM-adapter with those methods that are capable of processing more than two modalities simultaneously. 

\Cref{tab:DELIVER_all} shows the results on the DeLiVER validation and test sets. We employ the official weights of the competitors trained on all available modalities (RGB, LiDAR, Depth, and Event). We note that, even though trained on \emph{only two modalities}, our MM SAM-adapter achieves \textit{state-of-the-art} performance on the DeLiVER \cite{zhang2023delivering} test set regardless of the adopted auxiliary modality, and the best results also on the validation set when processing RGB and Depth.
\interfootnotelinepenalty=10000
\Cref{tab:MUSES_all} reports results on the MUSES dataset. Again,  MM SAM-adapter achieves \textit{state-of-the-art} performance independently of the adopted auxiliary modality, i.e. with both RGB + Event as well as  RGB + LiDAR, highlighting the potential of our approach for broad applicability to real contexts. \textit{These results are publicly available in the MUSES online benchmark}
 \footnote{\label{footnote}“MUSES online benchmark,” May 2025: \url{https://muses.vision.ee.ethz.ch/benchmarks\#semanticSegmentation}}.
\begin{figure*}[!h]
    \centering
    \includegraphics[width=0.77\linewidth]{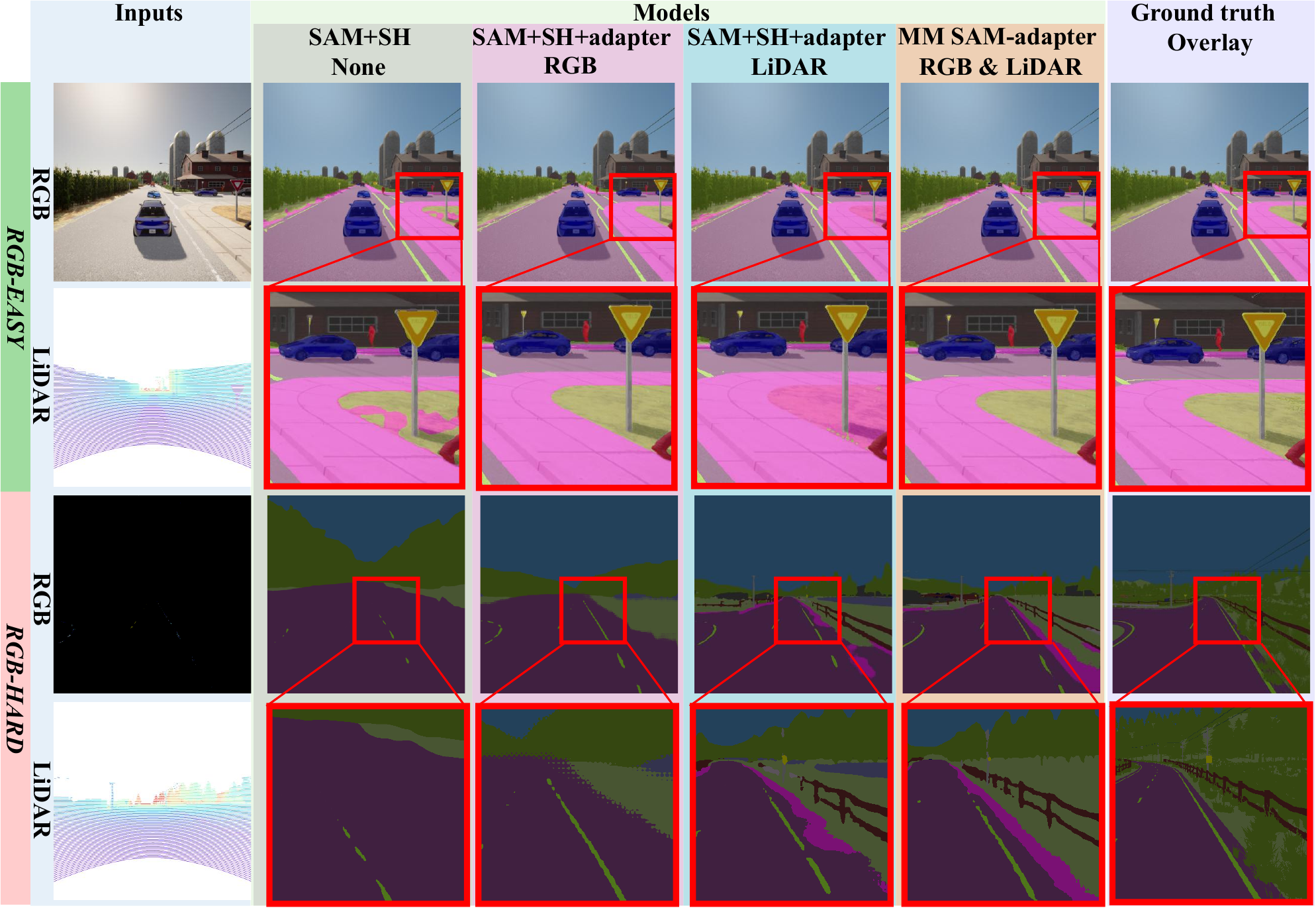}
    \caption{\textbf{DeLiVER test set predictions of some of the methods in \Cref{tab:DELIVER_ablated}} in RGB-LiDAR framework. \textit{RGB-easy}  (top) and \textit{RGB-hard} (bottom) samples.}
    \label{fig:qualitatives_ablations_}
\end{figure*}
\hl{Therefore, the experiments in this section confirm that our approach—based on adapting the SAM backbone with \textit{multimodal fused features}—is capable of effectively leveraging auxiliary knowledge while preserving the strong priors derived from the RGB-pretrained encoder.
}
\subsection{Ablation studies} \label{subsec:Ablation} 

\subsubsection{Main Contributions}
We conduct experiments on DeLiVER to highlight the impact of the key contributions in our proposed multimodal adapter architecture, reporting results in \Cref{tab:DELIVER_ablated} and \Cref{fig:qualitatives_ablations_}.
\begin{table}[t]
\caption{\textbf{Analysis of our contributions on the DeLiVER test set}. SH = Segmentation Head, SPM = Spatial Prior Module. SAM* = SAM backbone frozen}
    \centering
    \resizebox{\linewidth}{!}{
    \begin{tabular}{l|c|c|c|c|c}
        \toprule
        \multirow{2}{*}{\textbf{Method}} 
        & \textbf{Auxiliary} & \textbf{adapter} & \cellcolor{cyan20}\textbf{All} & \cellcolor{PeaGreen}\textbf{RGB-easy} & \cellcolor{red20}\textbf{RGB-hard} \\
        & \textbf{Encoders} & \textbf{Modalities} & \textbf{mIoU} & \textbf{mIoU} & \textbf{mIoU} \\
        \midrule
        \cellcolor{none} SAM + SH & None & None & 53.32 & 54.28 & 35.57 \\
         
        \cellcolor{white} SAM + SH + adapter & 1 SPM & RGB & 55.90 & 56.93 & 37.13 \\
       
        \cellcolor{samrgb} SAM* + SH + LoRA & 1 ConvNext-S & RGB & 53.18 & 54.28 &  35.05\\
        \cellcolor{samrgb} SAM + SH + adapter & 1 ConvNext-S & RGB & 55.98 & 57.07 & 37.93 \\
        \cellcolor{samlidar} SAM* + SH + LoRA & 1 ConvNext-S & LiDAR & 49.77 & 50.81 & 32.59 \\
        \cellcolor{samlidar} SAM + SH + adapter & 1 ConvNext-S & LiDAR & 54.90 & 55.49 & 44.06 \\
        
        \cellcolor{samrgblidar} MM SAM*-LoRA & 2 ConvNext-S & RGB \&  LiDAR & 53.97 & 54.74 & 40.22 \\
        \cellcolor{samrgblidar} \textbf{MM SAM-adapter} & 2 ConvNext-S & RGB \&  LiDAR & \textbf{57.14} & \textbf{57.75} & \textbf{45.46} \\
        
        \bottomrule
    \end{tabular}}
    
    \label{tab:DELIVER_ablated}
\end{table}
The first row shows the results of the SAM image encoder fine-tuned for semantic segmentation by simply appending a SegFormer \cite{xie2021segformer} Segmentation Head. The second and fourth rows, instead, report the results achieved by fine-tuning the SAM encoder via adapter-based approaches that utilize two different backbones: a Spatial Prior Module (SPM), as proposed in ViT-adapter \cite{chen2023vision}, and a standard ConvNext-Small (ConvNext-S). In these two experiments, the adapter branch takes only RGB images as input, akin to the original ViT-adapter. We observe that the side-tuning SAM-adapter outperforms standard fine-tuning, likely because it better preserves SAM's prior knowledge, reducing catastrophic forgetting.
\hl{The third row depicts the result of SAM image encoder frozen, adapted using a different adaptation strategy based on LoRA}\cite{hu2022lora}. \hl{To ensure a fair comparison with the SAM-adapter, we employ a ConvNext-S before the first LoRA layer as explained in Appendix} \ref{sec:LoRA}.
\hl{When comparing the second, third, and fourth rows, we observe that the side-tuning SAM-adapter surpasses the LoRA adaptation, likely due to its ability to smoothly inject new domain-specific spatial information.}.
However, RGB-only adapters still struggle in the \textit{RGB-hard} scenario, with a performance gap of about 20\% mIoU vs. \textit{RGB-easy}. \hl{In the fifth row, 
SAM with LoRA adapter performs poorly when the LiDAR modality is introduced at the first layer and unexpectedly struggles even in the} \textit{RGB-hard} \hl{scenario (from 50.81 in} \textit{RGB-easy} \hl{to 32.59 mIoU in} \textit{RGB-hard}).\hl{ This is likely because a single LoRA layer is insufficient to integrate a modality like LiDAR into the RGB-based SAM backbone, given the inherent differences between RGB and LiDAR data.} In the sixth row, we demonstrate that replacing the RGB input with the LiDAR modality in the \hl{side-tuning} adapter branch significantly improves performance in the \textit{RGB-hard} scenario (e.g., 44.06 vs. 37.93  mIoU for  LiDAR vs. RGB adapter with the same ConvNext-S backbone). However, this comes at the cost of a performance drop in the \textit{RGB-easy} scenario (from 57.07 to 55.49 mIoU),  as noisier LiDAR measurements may corrupt highly informative RGB features. \hl{The seventh row shows the results of adapting using LoRA, yet by employing multimodal data as input. Multimodal LoRA utilizes our MM Fusion Encoder, which comprises two modality-specific encoders and a Fusion module. The fused encoder features are fed into the first LoRA layer. 
The results align with our main idea that we need to inject multimodal information to achieve effective adaptation.  Indeed, multimodal LoRA obtains better performance than its single-modality counterpart in both \textit{RGB-easy} and \textit{RGB-hard} scenarios. However, we highlight that the best results are achieved by our proposed MM SAM-adapter strategy, as shown in the last row.} Notably, our approach achieves the best performance across all scenarios. 
\Cref{fig:qualitatives_ablations_} shows the predictions obtained by some of the methods in \Cref{tab:DELIVER_ablated}. Our MM SAM adapter yields way more accurate predictions than the RGB-only counterparts in the \textit{RGB-hard} scenario. In the \textit{RGB-easy} scenario, our model provides very similar segmentations compared to the RGB-only adapter, while performing much better than the model that uses the LiDAR-only auxiliary branch. These qualitative results vouch for our framework's ability to learn which modality to prioritize during inference, namely to discard or deploy LiDAR features depending on whether the RGB content is sufficiently informative or not. \hl{This capability is achieved through the combination of the Multimodal Fusion Encoder with the adapter module, which smoothly contaminates SAM foundational knowledge with multimodal one.}


\subsubsection{Asymmetric vs Symmetric Architecture}\label{subsec:ablation_symm}
To validate our asymmetric network design,  we replace the two ConvNext-Small (ConvNeXt-S) encoders in the auxiliary branch with two ConvNext-Base (ConvNeXt-B) networks, so as to realize a symmetric architecture where the auxiliary branch (Multimodal fusion encoder + adapter) and the SAM encoder have a similar number of parameters. In \Cref{tab:DELIVER_ablated_conv}, we present results on the DeLiVER test set for the RGB-LiDAR setup. Interestingly, the symmetric  MM SAM-adapter performs worse, which underscores the importance of maintaining an asymmetric structure to prioritize SAM knowledge over the auxiliary modality one. Additionally, our asymmetric design enables to use lighter networks for the auxiliary branch, reducing computational complexity without sacrificing performance. \hl{This experiment aligns with our intuition that RGB is typically the primary source of information, while auxiliary modalities are needed mainly when RGB data is lacking. Therefore, adopting a symmetric architecture could undermine this logic by assigning excessive emphasis to the auxiliary modality.}


\begin{table}[t]
    \centering
        \caption{\textbf{Symmetric vs asymmetric architecture.} Results on DeLiVER test set in the RGB-LiDAR setup.}
    \resizebox{\linewidth}{!}{
    \begin{tabular}{l|c|c|c|c}
        \toprule
        \multirow{2}{*}{\textbf{Method}}
        &\multirow{2}{*}{\textbf{Architecture}} & \cellcolor{cyan20}\textbf{All} & \cellcolor{PeaGreen}\textbf{RGB-easy} & \cellcolor{red20}\textbf{RGB-hard} \\
         &  & \textbf{mIoU} & \textbf{mIoU} & \textbf{mIoU} \\
        \midrule
        MM SAM-adapter & Symmetric & 55.27 & 55.93 & 42.22 \\
        \textbf{MM SAM-adapter} & Asymmetric & \textbf{57.14} & \textbf{57.75} & \textbf{45.46} \\
        \bottomrule
    \end{tabular}}

    \label{tab:DELIVER_ablated_conv}
\end{table}

\begin{table}[t]
    \centering
    \caption{\textbf{Comparison of fusion modules.} Results on the DeLiVER test in the RGB-LiDAR setup. \textit{Addition} or \textit{Concatenation} sums or concatenates the two modality features, respectively. \textit{Road-fusion} is the RoadFormer+ \cite{huang2024roadformer+} fusion module.}
    \resizebox{\linewidth}{!}{
    \begin{tabular}{l|c|c|c|c}
        \toprule
        \multirow{2}{*}{\textbf{Method}}
        & \textbf{Fusion} & \cellcolor{cyan20}\textbf{All} & \cellcolor{PeaGreen}\textbf{RGB-easy} & \cellcolor{red20}\textbf{RGB-hard} \\
         & \textbf{Module} & \textbf{mIoU} & \textbf{mIoU} & \textbf{mIoU} \\
        \midrule
        MM SAM-adapter & Addition & 56.09 & 56.66 & \textbf{45.53} \\
        MM SAM-adapter & Concatenation & 56.87 & 57.51 & 44.78 \\
        \textbf{MM SAM-adapter} & Road-Fusion & \textbf{57.14} & \textbf{57.75} & 45.46 \\
        \bottomrule
    \end{tabular}}
    
    \label{tab:DELIVER_ablated_fusion}
\end{table}

\subsubsection{Alternative Fusion Modules}\label{subsec:ablation_fusion}
\hl{Our architecture is flexible, as it can incorporate different fusion modules. Even the simpler fusion strategies enable the model to achieve remarkable performance.}
In ~\Cref{tab:DELIVER_ablated_fusion}, we explore alternative architectures for the Fusion Module. In particular, we replace  Road-Fusion \cite{huang2024roadformer+} with simpler fusion techniques such as addition and concatenation.
First of all, we point out that our intuition of adapting SAM using fused multimodal features proves effective with all the considered fusion techniques: even simple ones, like addition and  concatenation, can yield \textit{state-of-the-art} results on the RGB-LiDAR setup of DeLiVER. 
We also notice that in the case of addition-based fusion, the results for \textit{RGB-hard} samples are comparable to those of Road-Fusion. In contrast, a drop in performance is observed in the \textit{RGB-easy} case. This aligns with our expectations, as addition-based fusion makes it harder for the network to determine what information to use based on the scenario due to LiDAR noise being directly injected in SAM features. Conversely, with concatenation-based fusion, the results are close to those of Road-Fusion, especially in the \textit{RGB-easy} scenario. Indeed, concatenating the features from the two modalities allows the adapter to easily sift out only the required information at test time. Nevertheless, the Road-Fusion strategy generates better fused features, enabling the adapter to inject and extract knowledge effectively, thereby achieving superior performance in both \textit{RGB-easy} and \textit{RGB-hard} scenarios. 


\begin{table}[t]
    \centering
    \caption{Single modality-agnostic vs two modality-specific encoders. Results on the DeLiVER test set in the RGB-LiDAR setup.  \\ MA = Modality-Agnostic. MS = Modality-Specific.}
    \resizebox{\linewidth}{!}{
    \begin{tabular}{l|c|c|c|c}
        \toprule
        \multirow{2}{*}{\textbf{Method}}
        & \textbf{Auxiliary} & \cellcolor{cyan20}\textbf{All} & \cellcolor{PeaGreen}\textbf{RGB-easy} & \cellcolor{red20}\textbf{RGB-hard} \\
         & \textbf{Encoders} & \textbf{mIoU} & \textbf{mIoU} & \textbf{mIoU} \\
        \midrule
        MM SAM-adapter & MA & 56.21 & 56.81 & 44.79 \\
        \textbf{MM SAM-adapter} & MS & \textbf{57.14} & \textbf{57.75} & \textbf{45.46} \\
        \bottomrule
    \end{tabular}}
    
    \label{tab:DELIVER_ablated_shared}
\end{table}
\begin{table}[t]
    \centering
    \caption{\textbf{Comparison of our method vs SAM-frozen version.} Results on DeLiVER test set in the RGB-LiDAR setup.}
    \resizebox{\linewidth}{!}{
    \begin{tabular}{l|c|c|c|c}
        \toprule
        \multirow{2}{*}{\textbf{Method}} & \multirow{2}{*}{\textbf{SAM frozen}}& \cellcolor{cyan20}\textbf{All} & \cellcolor{PeaGreen}\textbf{RGB-easy} & \cellcolor{red20}\textbf{RGB-hard} \\
         &  & \textbf{mIoU} & \textbf{mIoU} & \textbf{mIoU} \\
        \midrule
        MM SAM-adapter & \checkmark & 55.35 & 56.00 & 43.30 \\
        \textbf{MM SAM-adapter} & \ding{55} & \textbf{57.14} & \textbf{57.75} & \textbf{45.46} \\
        \bottomrule
    \end{tabular}}
    
    \label{tab:DELIVER_ablated_frozen}
\end{table}

\begin{table}[t]
    \centering
    \caption{\textbf{Comparison of our method vs scratch-trained ViT-L backbone version.} Results on DeLiVER test set in the RGB-LiDAR setup.\\  ST = scratch-training of ViT-L with no pretrained weights, FT = fine-tuning of SAM from pretrained weights.}
    \resizebox{\linewidth}{!}{
    \begin{tabular}{l|c|c|c|c}
        \toprule
        \multirow{2}{*}{\textbf{Method}}& \multirow{2}{*}{\textbf{SAM training}}& \cellcolor{cyan20}\textbf{All} & \cellcolor{PeaGreen}\textbf{RGB-easy} & \cellcolor{red20}\textbf{RGB-hard} \\
         & & \textbf{mIoU} & \textbf{mIoU} & \textbf{mIoU} \\
        \midrule
        MM SAM-adapter & ST & 53.73 & 54.46 & 40.97 \\
        \textbf{MM SAM-adapter} & FT & \textbf{57.14} & \textbf{57.75} & \textbf{45.46} \\
        \bottomrule
    \end{tabular}}
    
    \label{tab:DELIVER_ablated_fullvsfine}
\end{table}
\subsubsection{Modality Specific vs Modality Agnostic Encoders}
To assess whether using two modality-specific encoders in the Multimodal Fusion Encoder is more effective than a shared, modality-agnostic encoder, we carry out an ablation study on the DeLiVER \cite{zhang2023delivering} test set in the RGB-LiDAR setup. The findings, shown in  \Cref{tab:DELIVER_ablated_shared}, indicate that employing two modality-specific encoders performs better than a single modality-agnostic encoder for the RGB-LiDAR modality pair. This can be attributed to the fundamental differences between RGB and LiDAR data: while RGB images provide dense visual information, LiDAR data is inherently sparse. 
\subsubsection{Frozen vs Finetuned SAM}
In~\Cref{tab:DELIVER_ablated_frozen}, we compare our method with a variant where the SAM image backbone remains frozen during training. As noted by ViT-adapter \cite{chen2023vision}, this adaptation strategy benefits significantly from fine-tuning the SAM backbone. Consequently, fine-tuning SAM achieves the best performance.
\hl{Remarkably, the results obtained with the SAM-frozen backbone are nearly the same as CAFuser (i.e., 55.35 vs. 55.60 mIoU -- see } \Cref{tab:DELIVER_all}).\hl{ This demonstrates that, even with the backbone entirely frozen, our multimodal adapter and fusion modules effectively guide the SAM backbone by integrating multimodal information in a beneficial way.}

\subsubsection{Importance of SAM pre-trained weights}
In~\Cref{tab:DELIVER_ablated_fullvsfine}, we present a reference baseline by evaluating our proposed architecture without using the pre-trained weights of the SAM image encoder. As expected, leveraging SAM’s pre-trained weights is essential for achieving \textit{state-of-the-art} performance. 

\section{Conclusion}\label{sec:conclusion}
We propose MM SAM-adapter, a novel multimodal semantic segmentation framework that leverages an adapter strategy to harness SAM’s rich foundational knowledge. By enabling the adapter to utilize fused features from the RGB and one auxiliary modality, MM SAM-adapter ensures robust performance in challenging conditions while maintaining optimal accuracy in simpler scenarios, resulting in state-of-the-art results across all the considered benchmarks.

One limitation of the MM SAM-adapter is that it currently supports only two input modalities due to constraints imposed by the road-fusion module. While our approach already outperforms existing all-modalities methods, an exciting avenue for future research is to extend this framework to accommodate more complex scenarios. This will require designing an innovative and efficient fusion module capable of effectively integrating more than two modalities. Additionally, exploring the potential of this framework in other tasks, such as panoptic segmentation, presents another promising avenue for further research.

\bibliographystyle{plain} 

\newpage
\section*{Biographies}

\noindent\textbf{Iacopo Curti}\quad
received his Master Degree in Automation Engineering in 2023 from the University of Bologna. He is a PhD student at the Department of Computer Science and Engineering, University of Bologna. His research interests include image processing and deep learning frameworks, in particular multimodal architectures for dense prediction tasks.

\vspace{1ex}
\noindent\textbf{Pierluigi Zama Ramirez}\quad
received his PhD in Computer Science and Engineering in 2021. He is currently an Assistant Professor at the University of Bologna. He co-authored more than 30 publications on computer vision research topics such as semantic segmentation, depth estimation, anomaly detection, domain adaptation, neural fields, and 3D computer vision.

\vspace{1ex}
\noindent\textbf{Alioscia Petrelli}\quad
received the Master Degree in Computer Science Engineering and the PhD Degree in Computer Science from the University of Bologna in 2005 and 2016, respectively. He spent four years as research fellow at the Computer Vision Laboratory (CVLab) of the Department of Computer Science and Engineering (DISI) in Bologna. His research interests focus on computer vision, including 3D surface matching, visual search, and machine learning.

\vspace{1ex}
\noindent\textbf{Luigi Di Stefano}\quad
received the PhD degree in electronic engineering and computer science from the University of Bologna in 1994. He is currently a full professor at the Department of Computer Science and Engineering, University of Bologna, where he founded and leads the Computer Vision Laboratory (CVLab). His research interests include image processing, computer vision, and machine/deep learning. He is the author of more than 150 papers and several patents. He has been a scientific consultant for major computer vision and machine learning companies. He is a member of the IEEE Computer Society and the IAPR-IC.

\clearpage
\newpage
\renewcommand{\thesection}{S}

\setcounter{figure}{0}
\setcounter{table}{0}
\renewcommand{\thefigure}{\Alph{figure}}
\renewcommand{\thetable}{\Alph{table}}
\section*{Additional Implementation Details}
\subsection{LoRA multimodal implementation}\label{sec:LoRA}
\hl{The LoRA} \cite{hu2022lora} \hl{multimodal network, showed in }\Cref{tab:DELIVER_ablated} \hl{has been implemented to assess the choice of the adapter method, presented in}\cite{chen2023vision}.\hl{ LoRA layers have been incorporated into every layer of the SAM Vision Transformer. However, unlike the standard LoRA implementation, we introduce modality-specific encoders and a fusion module at the first LoRA layer, to make it compatible with multimodal input. In the case of RGB-only data, a ConvNeXt network extracts features from the RGB image, which are then projected to a dimensionality compatible with SAM tokens. The standard LoRA layers subsequently process these projected features. For LiDAR input, the ConvNeXt network directly processes the LiDAR data, and the resulting features follow the same processing as in the RGB-only case. In the RGB-LiDAR scenario, the fusion module is identical to that used in the MM SAM-adapter, with two modality-specific encoders dedicated to processing RGB and LiDAR inputs, respectively.}
\subsection{Competitor details}
\hl{Each competitor has been trained and tested using the official repository provided by their authors. We used CMNeXt}\cite{zhang2023delivering} \href{https://github.com/jamycheung/DELIVER}{GitHub repository},   RoadFormer+\cite{huang2024roadformer+} \href{https://github.com/LiJiahang617/Road-Former}{GitHub repository}, GeminiFusion\cite{pmlr-v235-jia24b} \href{https://github.com/JiaDingCN/GeminiFusion}{GitHub repository}.
\section*{Additional qualitative results} \label{appendix}
\begin{figure}[h]
    \centering
    \includegraphics[width=0.9\linewidth]{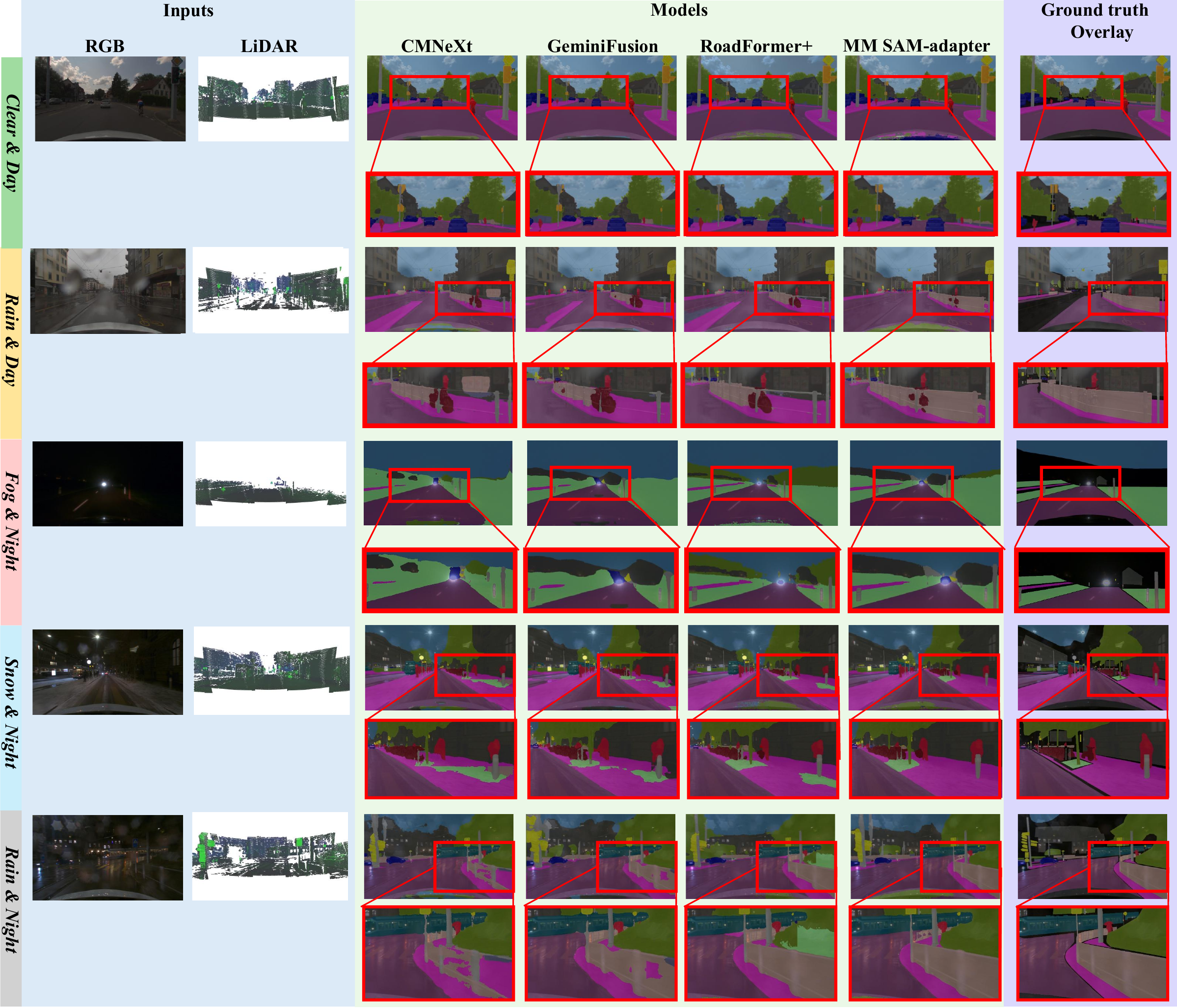}
    \caption{\textbf{MUSES \cite{brodermann2024muses} validation set predictions} in RGB-LiDAR framework. Different conditions have been illustrated.}
    \label{fig:qualitatives_MUSES_LIDAR}
\end{figure}

\begin{figure}[h]
    \centering
    \includegraphics[width=0.9\linewidth]{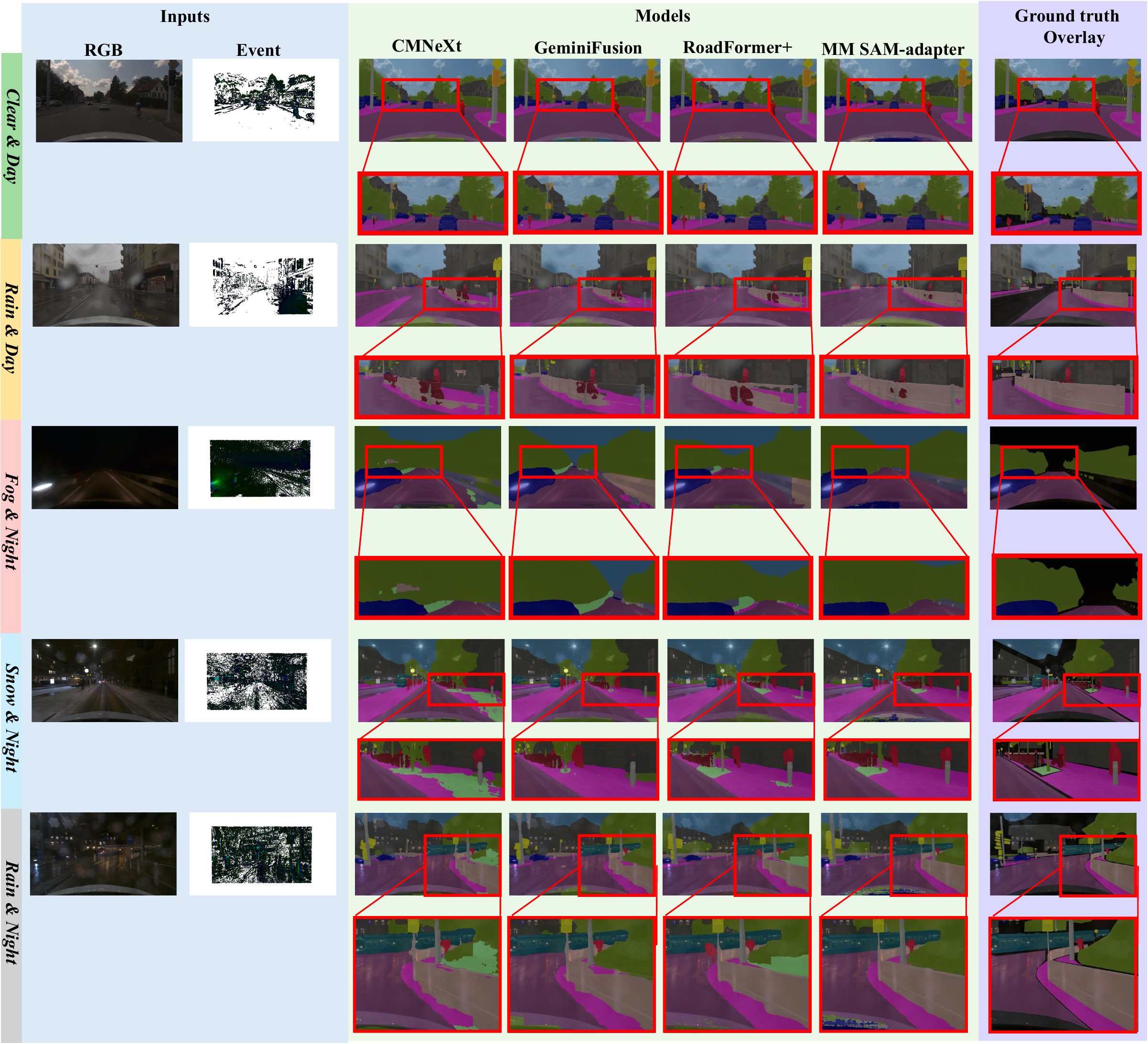}
    \caption{\textbf{MUSES \cite{brodermann2024muses} validation set predictions} in RGB-Event framework. Different conditions have been illustrated}
    \label{fig:qualitatives_MUSES_EVENT}
\end{figure}

\begin{figure}[t]
    \centering
    \includegraphics[width=0.86\linewidth]{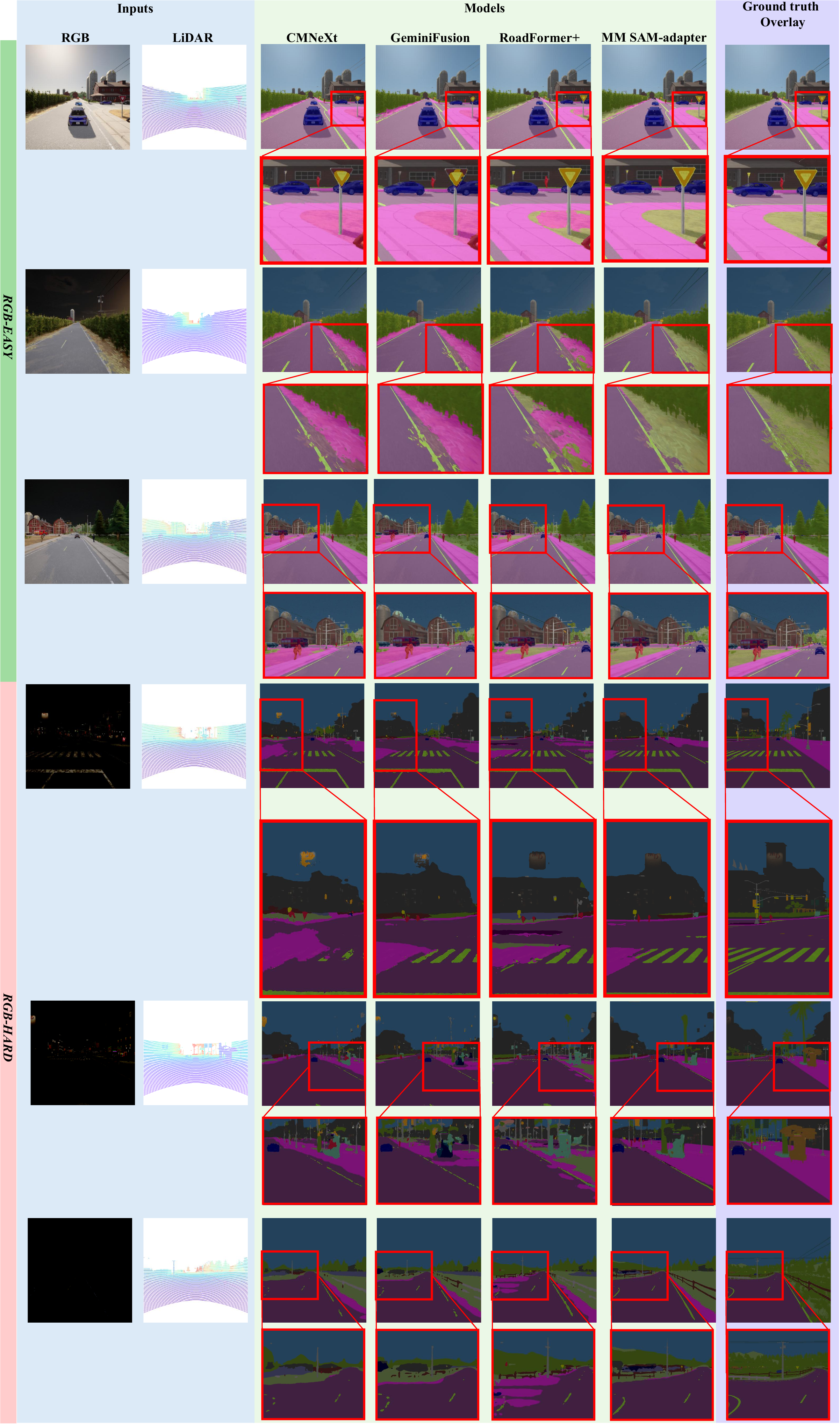}
    \caption{\textbf{DeLiVER \cite{zhang2023delivering} test set predictions} in RGB-LiDAR framework. \textit{RGB-easy}  (top) and \textit{RGB-hard} (bottom) samples. Notably, the input RGB image in \textit{RGB-hard} case has been stretched with an exponential operator.}
    \label{fig:qualitatives_LiDAR_add}
\end{figure}

\begin{figure}[!h]
    \centering
    \includegraphics[width=0.9\linewidth]{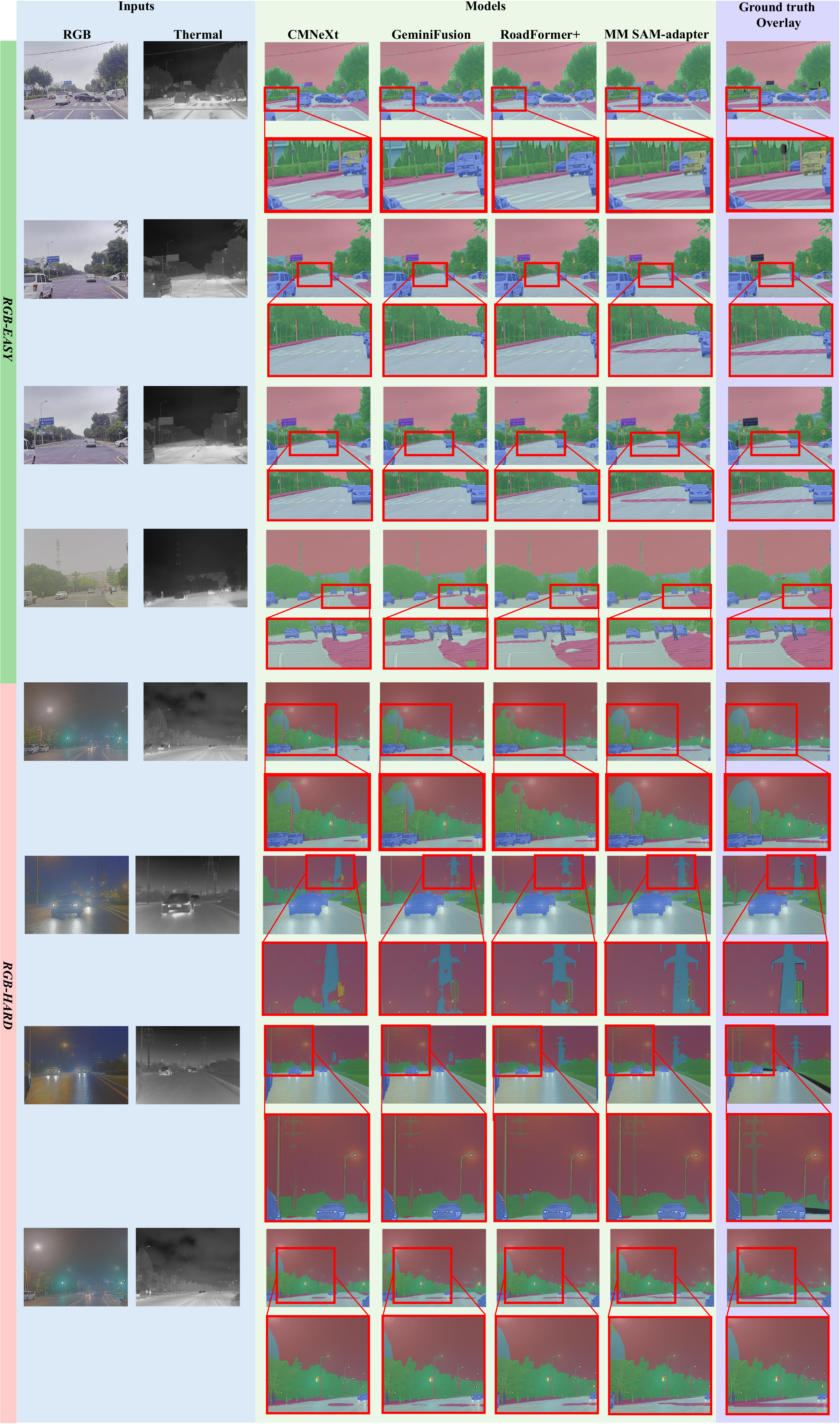}
    \caption{\textbf{FMB \cite{liu2023segmif} test set predictions}. \textit{RGB-easy}  (top) and \textit{RGB-hard} (bottom) samples.}
    \label{fig:qualitatives_FMB_add}
\end{figure}

We present qualitative results of our proposed method and competitors in \Cref{fig:qualitatives_MUSES_LIDAR}, \Cref{fig:qualitatives_MUSES_EVENT}, \Cref{fig:qualitatives_LiDAR_add}, \Cref{fig:qualitatives_FMB_add}.

\end{document}